\documentclass[a4paper]{IEEEtran}
\IEEEoverridecommandlockouts

\usepackage{graphicx}
\usepackage{multirow}
\usepackage{amsmath,amssymb,amsfonts}
\usepackage{amsthm}
\usepackage{mathrsfs}
\usepackage{xcolor}
\usepackage{textcomp}
\usepackage{manyfoot}

\usepackage{algorithm}
\usepackage{algorithmicx}
\usepackage{algpseudocode}
\usepackage{listings}

\usepackage{booktabs}
\usepackage{cite}
\usepackage{color}
\usepackage{soul}
\usepackage{hyperref} 
\usepackage{caption} 
\usepackage{subcaption} 
\usepackage{rotating} 
\usepackage{multirow} 
\usepackage{nicefrac} 
\usepackage{tikz} 
\usepackage{mathabx} 
\usepackage[normalem]{ulem} 
\usepackage{siunitx}
\usepackage{comment}
\usepackage{orcidlink}
\usepackage{adjustbox}  
\usepackage{cleveref}

\usepackage[accsupp]{axessibility}  

\usepackage{hyperref}

\usepackage{orcidlink}

\newcommand{\etal}{\textit{et al}.}

\begin{document}

\def\mao#1{\noindent\textcolor{red}{Mao: \textit{#1}}}
\def\lirfu#1{\noindent\textcolor{Green}{\fbox{lirfu} \textit{#1}}}
\def\petra#1{\noindent\textcolor{RubineRed}{Petra: \textit{#1}}}

\title{Sequential PatchCore: Anomaly Detection for Surface Inspection using Synthetic Impurities}

\author{
    \IEEEauthorblockN{\orcidlink{0009-0009-5230-4157} Runzhou Mao*},
    \IEEEauthorblockA{
        \textit{Fraunhofer ITWM},
        Kaiserslautern, Germany \\
    }
    \and
    \IEEEauthorblockN{\orcidlink{0000-0003-1858-5978} Juraj Fulir*},
    \IEEEauthorblockA{
        \textit{Fraunhofer ITWM},
        Kaiserslautern, Germany \\
    }
    \and[\hfill\mbox{}\par\mbox{}\hfill]
    \IEEEauthorblockN{\orcidlink{0000-0003-1669-8549} Christoph Garth},
    \IEEEauthorblockA{
        \textit{RPTU Kaiserslautern-Landau},
        Kaiserslautern, Germany \\
    }
    \IEEEauthorblockN{\orcidlink{0000-0002-2102-2618} Petra Gospodneti{ć}},
    \IEEEauthorblockA{
        \textit{Fraunhofer ITWM},
        Kaiserslautern, Germany \\
    }
}

\maketitle

\def\thefootnote{*}\footnotetext{Equal contribution}\def\thefootnote{\arabic{footnote}}

\begin{abstract}
    The appearance of surface impurities (e.g., water stains, fingerprints, stickers) is an often-mentioned issue that causes degradation of automated visual inspection systems. 
    At the same time, synthetic data generation techniques for visual surface inspection have focused primarily on generating perfect examples and defects, disregarding impurities.
    This study highlights the importance of considering impurities when generating synthetic data.
    We introduce a procedural method to include photorealistic water stains in synthetic data.
    The synthetic datasets are generated to correspond to real datasets and are further used to train an anomaly detection model and investigate the influence of water stains.
    The high-resolution images used for surface inspection lead to memory bottlenecks during anomaly detection training.
    To address this, we introduce Sequential PatchCore - a method to build coresets sequentially and make training on large images using consumer-grade hardware tractable.
    This allows us to perform transfer learning using coresets pre-trained on different dataset versions.
    Our results show the benefits of using synthetic data for pre-training an explicit coreset anomaly model and the extended performance benefits of finetuning the coreset using real data.
    We observed how the impurities and labelling ambiguity lower the model performance and have additionally reported the defect-wise recall to provide an industrially relevant perspective on model performance.
\end{abstract}
\begin{IEEEkeywords}
    synthetic impurities, anomaly detection, surface inspection
\end{IEEEkeywords}

\section{Introduction}
\label{sec:intro}
\begin{figure}[tb]
  \centering
  \begin{subfigure}[b]{0.32\linewidth}
    \centering
    \includegraphics[width=\linewidth]{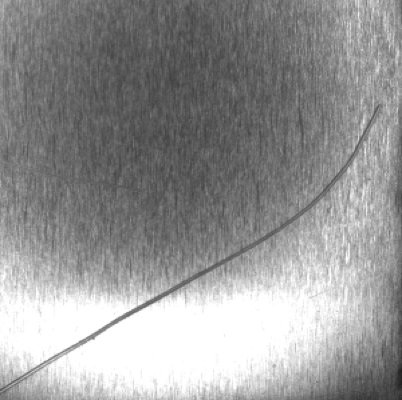}
    \caption{}
  \end{subfigure}
  \begin{subfigure}[b]{0.32\linewidth}
    \centering    
    \includegraphics[width=\linewidth]{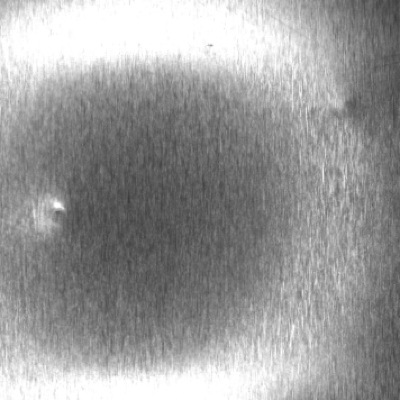}
    \caption{}
  \end{subfigure}
  \begin{subfigure}[b]{0.32\linewidth}
    \centering
    \includegraphics[width=\linewidth]{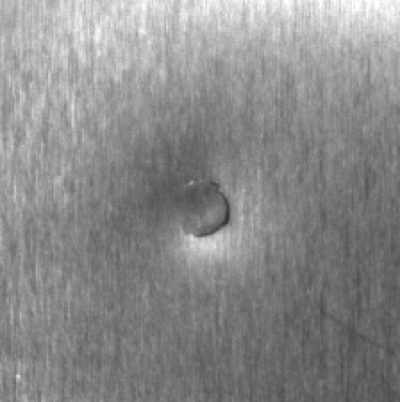}
    \caption{}
  \end{subfigure}
  \begin{subfigure}[b]{0.32\linewidth}
    \centering
    \includegraphics[width=\linewidth]{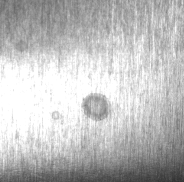}
    \caption{}
  \end{subfigure}
    \begin{subfigure}[b]{0.32\linewidth}
    \centering
    \includegraphics[width=\linewidth]{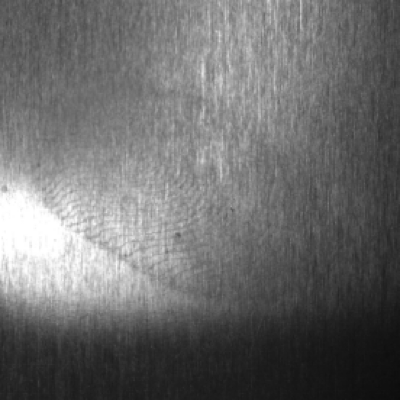}
    \caption{}
  \end{subfigure}
    \begin{subfigure}[b]{0.32\linewidth}
    \centering
    \includegraphics[width=\linewidth]{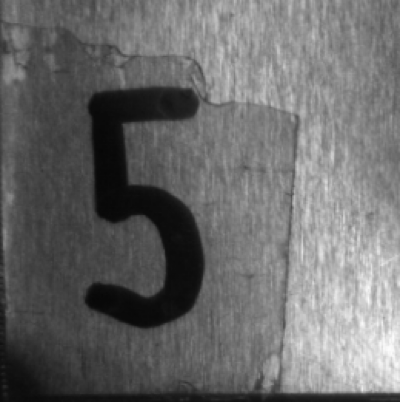}
    \caption{}
  \end{subfigure}
  
  \caption{Examples of surface defects and impurities in metal sheets. \textbf{Defects}: (a) scratch, (b) bump, (c) dent. \textbf{Impurities}: (d) water stain, (e) fingerprint, (f) sticker. The water stain depicted in (d) exhibits characteristics similar to the dent shown in (c), indicating potential challenges in distinguishing between them.}
  \label{fig:imperfections}
\end{figure}

Automated visual surface inspection is an important step in the manufacturing process, ensuring product quality and enabling operational efficiency.
The task of a surface inspection system is to consistently and robustly detect defects (e.g., scratches, bumps, dents).
However, no surface and no environment are perfect, which is reflected in frequent appearance of so-called surface impurities (e.g. water stains, fingerprints, stickers).
Their visual features often share similarities with defect features (see \cref{fig:imperfections}), causing either performance degradation of the algorithm, or making the development of the inspection system more complex.

What constitutes a defect is always defined by the customer commissioning the inspection system. Therefore, it is difficult to draw a strict line between surface defects and impurities.
For the purpose of this work, we consider defects to be geometrical imperfections in the surface structure \cite{SurfaceDefects}, while impurities refer to extraneous substances attached to the surface \cite{Kohli2018}.
This problem incentivizes the use of anomaly detection models, which can find different types of unobserved defects and could later be tuned to the target specification.

Recent advancements in computer vision have significantly improved surface inspection through deep learning models.
However, the issues with impurities have not disappeared \cite{Wagenstetter2024}.
Furthermore, the models rely heavily on large, diverse training datasets, which are often scarce and costly to annotate. 
Synthetic data generation \cite{Bosnar2024virtualinspection} has emerged as a promising solution to data scarcity, enabling automated generation of photorealistic images and precisely annotated masks.
However, synthetic data generation for surface inspection has primarily focused on defect simulation, neglecting the surface impurities.

In this work, we address these issues by using water stains to illustrate the importance of incorporating impurities into synthetic datasets.
Our contributions are as follows:
\begin{itemize}
    \item Novel method for generating synthetic water stains,
    \item Analysis over effects of impurities on anomaly detection models,
    \item Novel view on model performance for overcoming ambiguous labeling by using defect-wise recall,
    \item Novel method for resolving memory bottleneck of PatchCore \cite{Roth2022patchcore} anomaly detection for high resolution images,
    \item Dual datasets for anomaly detection on metal surface under the influence of impurities\footnote{Dataset URL: \url{https://fordatis.fraunhofer.de/handle/fordatis/412}}.
\end{itemize}

\section{Related Work}
\label{sec:related_work}

\subsection{Synthetic Image generation}

Research on synthetic image dataset generation can be divided into two main approaches: synthetic image generation using machine learning, where new images are generated based on what the model has observed; and rule-based generation using computer graphics to simulate new images\cite{Schladitz2023}. 
However, all of them are focusing on the generation of geometry, textures or defects and do not address the generation of surface impurities.
Furthermore, the visual surface inspection field lacks an analysis of the effects of impurities on recognition models, despite them being a known problem in industrial systems and shown to affect human inspectors and automatic systems alike \cite{Aust2021human_inspection_dirty,Ramalingam2019robot_inspection_stains,Daeschel2023food_inspection_human_residue}.

Generative models, such as Generative Adversarial Networks (GANs) \cite{Zhang2021,He2023,Wei2023} and Diffusion models \cite{Sun2024stablediffusion_anomaly}, were applied for the purpose of enriching industrial datasets through image generation.
However, generative models require a significant amount of data to produce realistic results.
They struggle to generate out-of-distribution samples, such as unseen perspectives or details, which limits their applicability in real scenarios. 
Additionally, generative models cannot ensure consistent lighting and spatial relationships within an image, which are crucial for precise surface inspection. 
For those reasons, our work focuses on rule-based generation, which provides a finer control when generating datasets.

Compared to generative methods, rule-based methods rely on physically based computer graphics and a well-defined set of rules to describe the inspection context \cite{Schladitz2023, Wagenstetter2024, Fulir2023, Moonen2023, Bosnar2020pipeline, Zhu2023synth_part_classify}. 
Moonen \etal \cite{Moonen2023} use rule-based methods to create photorealistic synthetic images of industrial scenes, which are primarily used for object detection and scene recognition. Bosnar \etal \cite{Bosnar2024virtualinspection}, on the other hand, concentrate on achieving the detail-oriented photorealism required for surface inspection and focus on higher level of control needed to generate high variations of surface textures and defects. 
The pipeline focuses on defining the inspection context, which includes a 3D model of the inspected object, the imaging setup (camera and light parameters) relative to the object, and surface properties (texture and defect parameters). 
Once the pipeline is established, generating diverse datasets that include both expected and out-of-distribution samples becomes straightforward by adjusting parameters.
This ensures contextual consistency and a balanced dataset content.
In this work, we extend the pipeline proposed by Bosnar \etal to include surface impurities.

While the topic of \textbf{impurity generation} has not yet been discussed in the context of synthetic data generation, it is a known concept within computer graphics (CG) community, often referred to by the term \textit{imperfection}.
Classical computer graphics methods typically focus on achieving visual realism for a single instance, which may not align with the needs of synthetic data generation for machine learning. 
Therefore, it is essential to generate an arbitrary number of instances and control their variation.
Impurity modeling can be observed from two perspectives: modeling a single impurity instance and the distribution of multiple impurities over the surface \cite{Becket1990}.

From the perspective of a single impurity instance, researchers typically focus on its properties and interactions with the surrounding environment, which is crucial for understanding and simulating various impurities.
For instance, generating rust requires understanding chemical reactions between metals and oxygen\cite{Ishikawa2016}, while generating textile stains requires knowledge of liquid diffusion in yarn\cite{Zheng2019}.
These methods, while accurate, are complex and costly, making them difficult to apply for large-scale synthetic data generation. 
To address these challenges, procedural methods using simplified physical models and noise functions can be employed to simulate these effects, thereby enhancing efficiency while maintaining a degree of realism.
For example, rust can be simulated using rule-based aggregation models, and stains can be created with fractal boundaries\cite{Becket1990}. This stain generation method is similar to the way liquid spreads on paper. However, our water stains are simulated on a metal surface, the boundaries are not as intricate.
Instead, we use Perlin noise \cite{Perlin1985} to simulate the appearance of water stains, leveraging its ability to create natural, continuous patterns that can be easily adjusted for different levels of detail and complexity.
These characteristics make it highly suitable for large-scale synthetic data generation.

The spatial distribution of impurities affect model training and performance by influencing data realism, feature learning, and the model's ability to generalize to real-world scenarios. 
The tendency distribution method based on surface curvature or air exposure can simulate environmental effects \cite{Wong1997, Schwärzler2007}, but it is computationally intensive and not suitable for all surfaces. 
For flat objects like metal plates, this method is overly complex and inefficient.
To better handle impurity placement, we can benefit from sampling methods. 
Random sampling is simple and fast but may result in uneven feature point distribution\cite{Bosch2007, Nord2014, Bosnar2023defects}. 
Poisson disk sampling achieves natural distribution but is computationally expensive\cite{Schladitz2023}. 
We adopts jittered sampling in this work, which randomly generates points within each cell, ensuring impurities are neither too clustered nor too evenly distributed. 
This distribution can increase data diversity, helping the ML model to better learn the features and patterns of different regions.

\subsection{Anomaly Segmentation for Surface Inspection}
Defect segmentation is commonly developed by fitting a model using annotated data \cite{Fulir2023, Wen2023steel_defdec_survey,Chen2021defect_detection_survey,Tulbure2022defect_detection_survey}.
However, defective samples and annotations can be expensive to acquire in large amounts and it can be difficult to describe all the defect types.
Thus, anomaly segmentation instead focuses only on a form of correct sample memorization and using this memory to grade how anomalous a defective patch is using a distance metric \cite{Liu2023anomaly_detection_survey,Liu2023dream,Roth2022patchcore,Y02020patchsvdd,Tao2022ad_industry_survey}.

We focus on a branch of anomaly detection methods which combines an image feature extractor with an explicit coreset for memorizing the nominal patterns as data points, usually differing in algorithms used to construct or use the coreset \cite{Cohen2020spadead,Defard2020padim,Roth2022patchcore,Gong2019neuralturingmemory,Zhang2023prototypicalresnet}.
Explicit coresets are modeled by using the datapoints to model a nominal zone in the feature space.
They are in contrast to methods using an implicit coreset formed through geometric constraints on the feature extractor \cite{Y02020patchsvdd} or learned mapping of the features with the boundary function \cite{Liu2023simplenet}.
Forms of explicit coresets have been used in combination with other methods to aid with different recognition and reconstrucion tasks \cite{Gong2019neuralturingmemory,Hou2021memorygan,Sinha2020smallgan_coreset}.

Patchcore \cite{Roth2022patchcore} uses internal layers of an ImageNet \cite{ImageNet} pre-trained feature extractor to construct feature vectors and a top-down coreset subsampling algorithm to reduce the coreset size which increases the model efficiency.
PaDiM \cite{Defard2020padim} and similar works \cite{Wan2022strongerpadim,Rippel2020gaussian_distro_ad,Li2022continual_ad} aim to compress the coreset by formulating it as a set of multivariate Gaussian distributions, which summarizes the information about the clusters of data.
Various approaches were proposed to enhance the speed of coreset reduction \cite{Saiku2022enhancepatchcore,li2023targetshooting,Donghyeong2023fapm}, which carries the most cost to the algorithm.
However, majority of the methods require large amounts of working memory to fit all of the image features and intermediate distance calculations, restricting their practical usage to small image and dataset sizes \cite{yao2024multiclassad,Yang2024textilead}.
In FAPM \cite{Donghyeong2023fapm}, authors split the coreset calculation into image regions to reduce the memory load.
In AnomalyDINO \cite{damm2024anomalydino} the coreset is built using only a few example images and foreground masked patches without additional compression.
In \cite{Yang2024textilead}, authors present a hybrid solution where the coreset is made out of image-wise features from a fixed number of random images from train-set, which are used by a student network via self-attention modulation in the reconstruction anomaly detection process.
In \cite{xie2023augmentedpatchcore} authors augment the training data to increase the data size, before applying coreset reduction.
In CFA \cite{Lee2022cfa} authors use k-means and exponential moving average to iteratively adapt it with new data.

Our work derives from PatchCore \cite{Roth2022patchcore} due to its simplicity of implementation and interpretation.
We reworked the original coreset reduction algorithm to a \textit{sequential} coreset building algorithm, which gives us more flexibility on limited-memory hardware and enables on-the-fly updating of the coreset with new data.

\section{Water Stain Generation}
\label{sec:water_stain}
\begin{figure}[tb]
  \centering
  \begin{minipage}{0.24\linewidth}
    \centering
    \includegraphics[width=\linewidth]{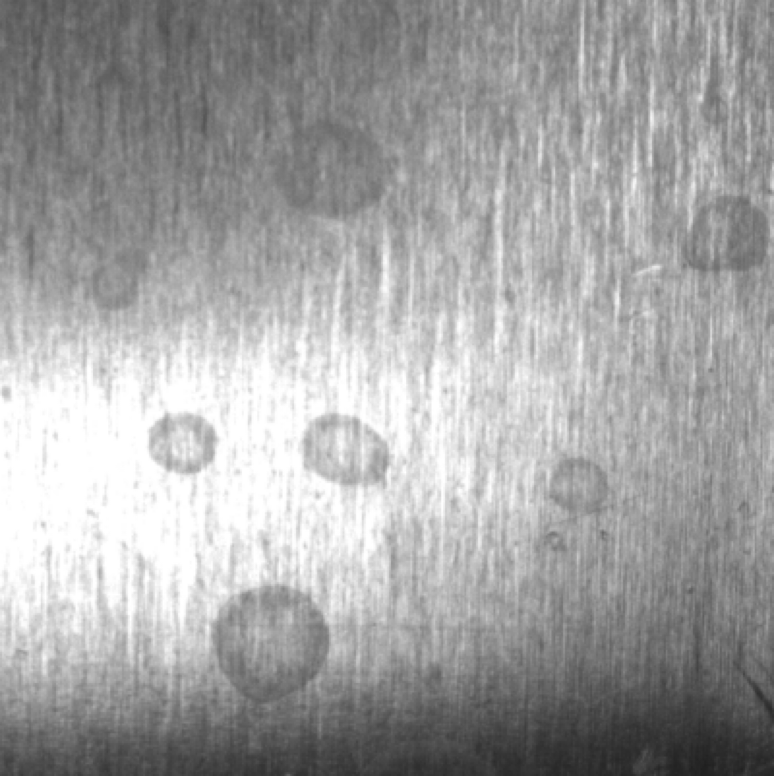}
  \end{minipage}
  \begin{minipage}{0.24\linewidth}
    \centering
    \includegraphics[width=\linewidth]{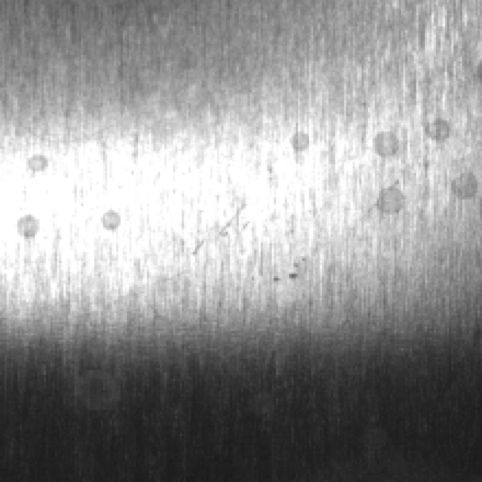}
  \end{minipage}
  \begin{minipage}{0.24\linewidth}
    \centering
    \includegraphics[width=\linewidth]{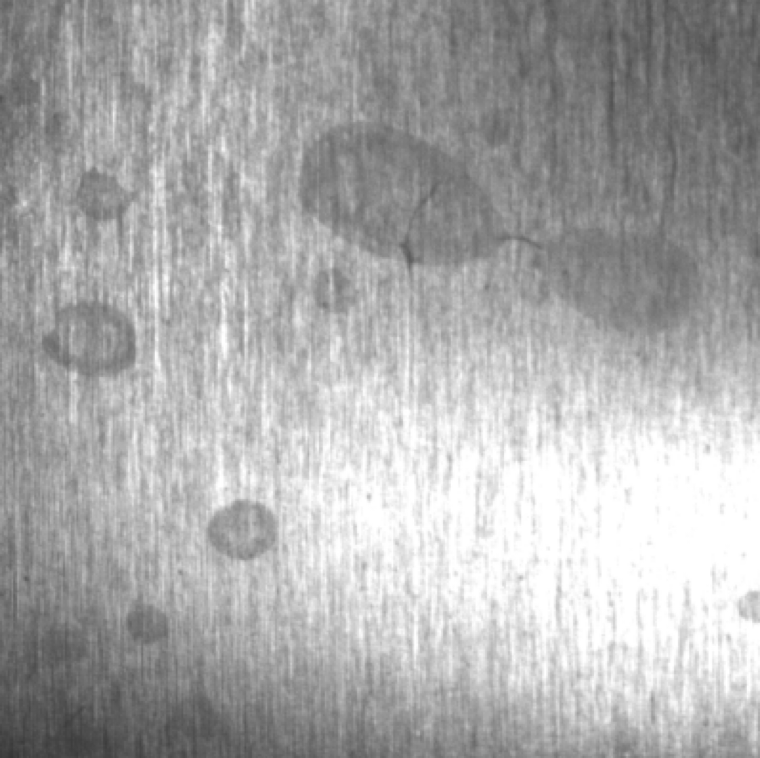}
  \end{minipage}
  \begin{minipage}{0.24\linewidth}
    \centering
    \includegraphics[width=\linewidth]{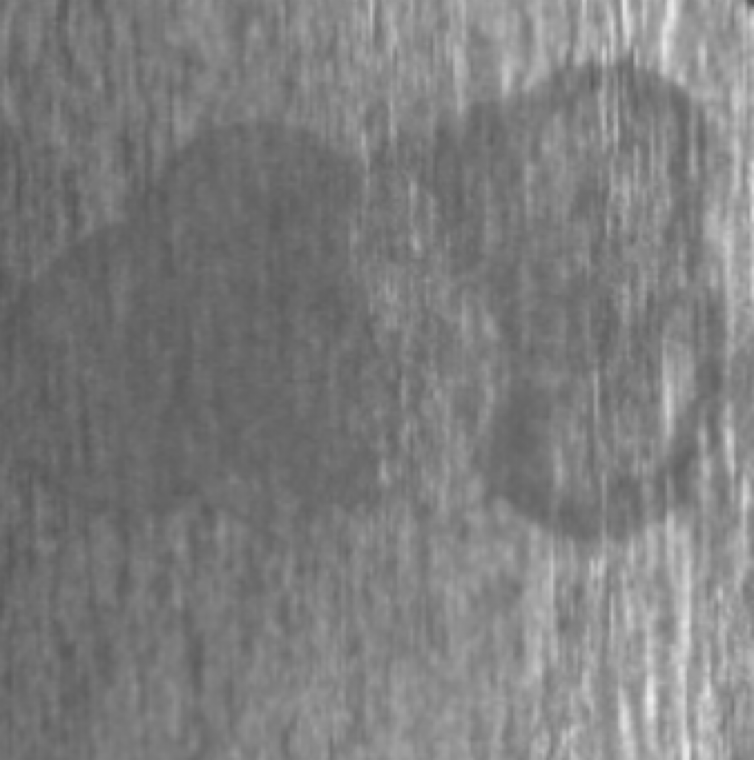}
  \end{minipage}
  
  \begin{minipage}{0.24\linewidth}
    \centering
    \includegraphics[width=\linewidth]{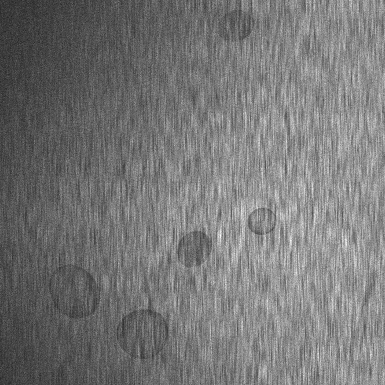}
  \end{minipage}
  \begin{minipage}{0.24\linewidth}
    \centering
    \includegraphics[width=\linewidth]{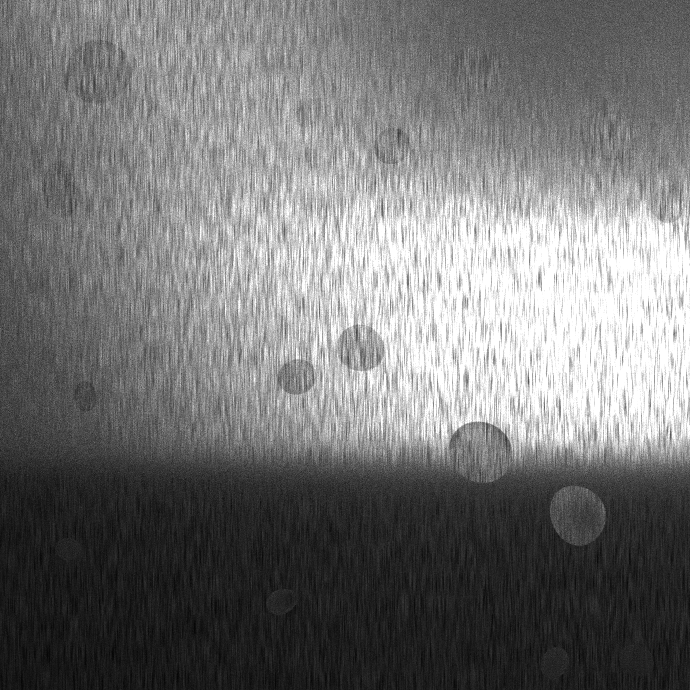}
  \end{minipage}
  \begin{minipage}{0.24\linewidth}
    \centering
    \includegraphics[width=\linewidth]{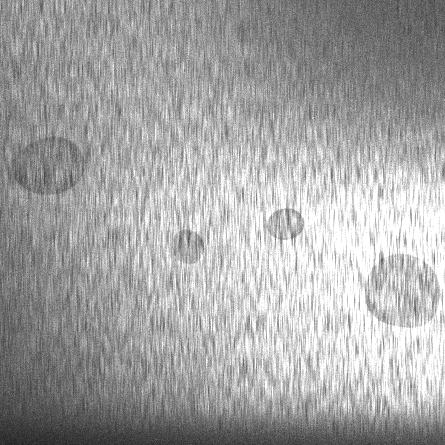}
  \end{minipage}
  \begin{minipage}{0.24\linewidth}
    \centering
    \includegraphics[width=\linewidth]{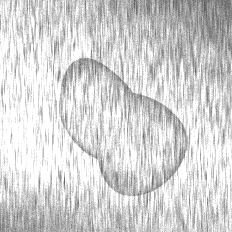}
  \end{minipage}
  
  \begin{minipage}{0.24\linewidth}
    \centering
    \includegraphics[width=\linewidth]{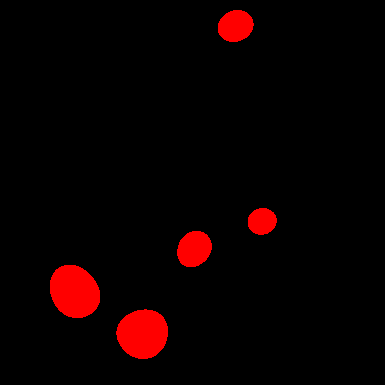}
  \end{minipage}
  \begin{minipage}{0.24\linewidth}
    \centering
    \includegraphics[width=\linewidth]{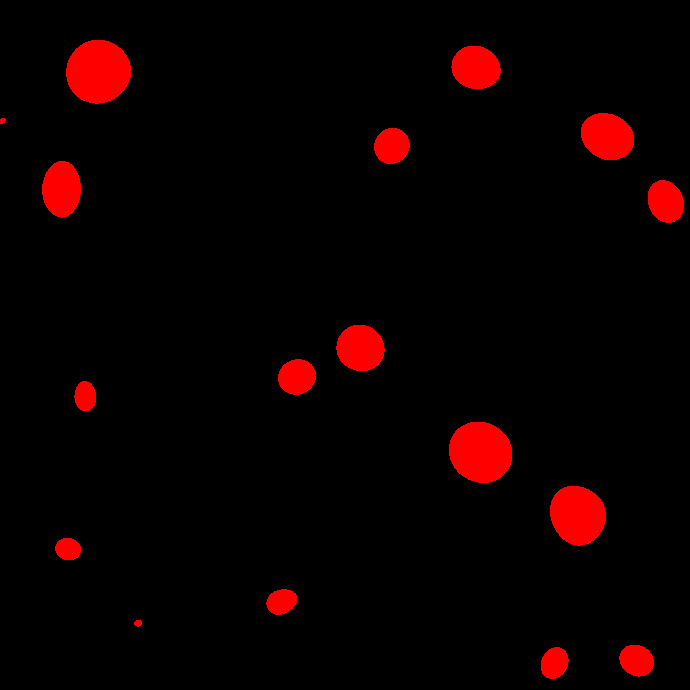}
  \end{minipage}
  \begin{minipage}{0.24\linewidth}
    \centering
    \includegraphics[width=\linewidth]{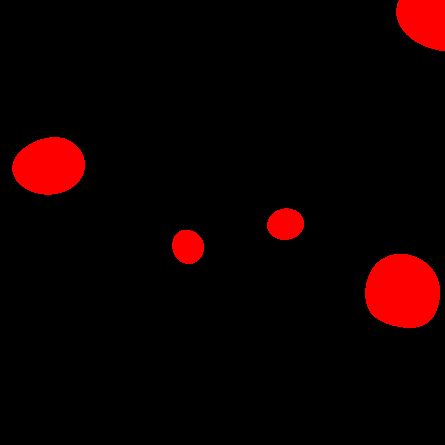}
  \end{minipage}
  \begin{minipage}{0.24\linewidth}
    \centering
    \includegraphics[width=\linewidth]{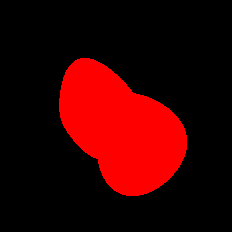}
  \end{minipage}
  \caption{Water stains of different shapes and sizes. Real water stain images (top); synthetic water stain images (middle); water stain segmentation masks (bottom).}
  \label{fig:water_stains}
\end{figure}

For the purpose of this work, we consider only water stains as they are the most common impurity in industrial environments. 
A water stain shape varies from circular to an irregular shape with a rounded outline (see \cref{fig:water_stains}).
Within the outline, one can notice a change in reflectance from the edges toward the center.
The central part of the stain is often close to the original surface reflectance, while the edges may appear darker.

To create realistic and varied water stains on object surfaces, we use solid texturing techniques.
Solid texturing is a process where a texture generation function is evaluated at every visible surface point of a model. 
As a result, the properties such as color, reflectance, or normals of a particular surface point depend solely on its three-dimensional position\cite{Ebert2002}.

The shape of a single water stain is modeled by starting from what we consider to be the basic shape of a water stain - a circle with a center $c$ and a radius $r$. 
To make the morphology of the stain more irregular, rather than having perfectly rounded boundaries, the edges of the circle are perturbed using Perlin noise \cite{Perlin1985}.  
Perlin noise is a gradient noise function that generates smooth, continuous variations, making it ideal for simulating these natural irregularities. 
To enable consistent evaluation within a single water stain, we use a predefined frequency $f$ and the current surface point $p$ as the seed values for the Perlin noise.
The final perturbation value is obtained by scaling the noise amplitude $A$.

To further enhance the realism, we simulate the optical properties of water stains — higher reflectance at the center and lower at the edges.
For each surface point $p$ within the water stain outline, we use its normalized distance $\hat{d}$ from the center point in order to determine the value of the exponential reflectance decay.
The final output value of the surface point ${R_o}_p$ is a combination of the underlying surface reflectance ${R_i}_p$ and the computed water stain reflectance.
The decay speed and water stain intensity are controlled by $\gamma$ and $\alpha$, respectively.

During rendering, for each point $p$ on the surface of the object, it is necessary to determine whether it is a part of a water stain or not.
For that we need to define the distribution of water stain center points.
To distribute multiple water stains across the object surface, we adopt the jittered sampling\cite{Nord2014}.
The three-dimensional space is sub-divided into grids, with a predetermined cell size $G$, and a single point $c_i$ within each cell $C_i$ is randomly generated to represent the water stain center.
The grid subdivision also allows us to optimize the calculation of the closest water stain, by evaluating only the distance to the center points within its own cell and immediate neighbor cells. 
By comparing the distance $d$ to the closest water stain with the perturbed radius $r'$, we determine whether the point lies within the water stain, and assign reflectance accordingly (see \cref{alg:water_stains}).
This process ensures that water stains are moderately dispersed across the surface and their density across the surface can be controlled by adjusting the grid cell size $G$. 

The same algorithm is used to generate pixel-precise annotations of the water stains (see \cref{fig:water_stains}).
Rather than calculating the reflectance change of $R_i$, a specific color is assigned to the area within the water stain, and black for everything else.

\begin{algorithm}[H]
    \caption{Water Stain Generation}
    \label{alg:water_stains}
    \begin{algorithmic}[1]
        \State \textbf{Input:} Surface point coordinate $p$, Radius $r$, Center $c$, Perlin frequency $f$, Perlin amplitude $A$, Input reflectance ${R_i}_p$, Reflectance decay $\gamma$, Reflectance intensity $\alpha$, Cell size $G$
        \State \textbf{Output:} Output reflectance ${R_o}_p$
        \State $r' \gets r + \text{PerlinNoise}(p, f) \times A$
        \If{$c$ is provided} 
            \State $d \gets \| p - c \| $
        \ElsIf{$G$ is provided}
            \State $c_i \gets \text{RandomPointInGridCell}(C_i) \quad \text{where} \quad C_i = \{p \mid i =  \lfloor \frac{p}{G} \rfloor\}$
            \State $d \gets \min \| p - c_i \| \quad \text{where} \quad c_i \in \text{neighbors}(C_i)$
        \EndIf
        \If{$d \leq r'$ }
            \State $\hat{d}\gets \frac{d}{r'}$
            \State ${R_o}_p \gets {R_i}_p + \text{pow}(\hat{d}, \gamma) \times \alpha$
        \Else
            \State ${R_o}_p \gets {R_i}_p $ 
        \EndIf
    \end{algorithmic}
\end{algorithm}

\section{Sequential PatchCore}
\label{sec:patch_core}

PatchCore \cite{Roth2022patchcore} consists of a parameterized patch feature extractor and a coreset memory bank $\mathcal{M}$.
The feature extractor extracts feature maps consisting of feature vectors describing each of the patches within an image.
The feature maps are average-pooled, upsampled to have matching size and concatenated along the channel dimension.
Average pooling increases the receptive field (patch size) without the need for features from the deeper layers, which would increase feature-space bias and reduce prediction resolution.
The coreset is a set of feature vectors, limited in size and collected from the nominal patches. 
It is used to measure the distance of tested patches from the region of nominal patches.
First the feature vectors of all the patches within the training set are collected using a feature extractor with parameters fixed at the values obtained from pre-training on ImageNet \cite{ImageNet}.
Then, the coreset is reduced to the specified size by iteratively finding a vector with the smallest distance to its closest neighbor and removing it from the coreset.
The goal of coreset reduction is to maximize the area of the nominal zone and uniformly cover it using the predefined number of points.

During inference, it evaluates each image patch by measuring its feature distance from the coreset and uses this distance to determine the level of anomaly.
The anomaly map is created by reassembling the patch anomaly levels, upscaling them using nearest-neighbor interpolation to match the original image size, and applying a blur to smooth out the blocky predictions.

The coreset collection and reduction process becomes prohibitively expensive for large datasets with high-resolution images, leading to memory overflows on consumer hardware.
The problem is that the feature vectors of all the patches in every image ($P$) across the entire dataset ($N$) must be collected before being reduced to the coreset size ($|\mathcal{M}|$).
Our goal is to reformulate the top-down coreset reduction algorithm to hold only the coreset in memory at any time, reducing its complexity from $O(NP)$ to $O(|\mathcal{M}| \ll NP)$.

Our \textit{sequential} algorithm starts with an empty coreset and iteratively adds patch features until the coreset achieves a predefined size.
Once the coreset is full, we calculate a dense distance matrix $D_\mathcal{M}$ between the samples in the coreset.
This structure is expensive to store, but we chose it to simplify implementation and speed up the search for nearest neighbors during training and inference. 
For every tested patch $p$, we find its closest coreset point $\mathcal{M}_p$ and their distance $d_p=d(p,\mathcal{M}_p)$.
The distance $d_p$ is compared to the smallest recorded distance $d_m=\min_{i,j} D_{\mathcal{M};i,j}$ within the coreset distance matrix.
If $d_p > d_m$, the tested patch replaces the one of the coreset patches $(\mathcal{M}_i,\mathcal{M}_j)$, with an implementation bias towards replacing the patch at the lower-index $min(i,j)$.
In this manner, the coreset keeps expanding the nominal zone in feature-space, while simultaneously ensuring a uniform coverage of the space within the zone.
Since the data is fed to the model in sequential order and its ordering affects the final coreset composition, the algorithm does not guarantee an optimal solution in a single epoch.
Therefore, we train our models for multiple epochs with early stopping when an epoch does not produce any additional changes to the coreset.
The sequential formulation also allows us to use data augmentation over multiple epochs in a memory efficient manner and consistent with traditional supervised learning pipelines, which was not possible with the original top-down approach.
Data augmentation will further increase the coverage of the coreset and help resolve misalignment issues noted in \cite{yao2024multiclassad}.
During inference we extract the patches of the input image, but we subdivide them into smaller chunks for which the distances from neighbors are calculated.
This allows us to manage the memory intensity of distance calculation (at the cost of runtime), while leveraging the GPU parallelization, which in turn results in overall speed increase.

Once we construct the dataset coreset, we can increase its reach by using the same algorithm to meld it with coresets collected from other datasets.
\textit{Coreset melding} is very fast since it skips the feature extraction and iteration over a large dataset, and instead iterates over the significantly smaller coresets.
This formulates a form of transfer learning, where the knowledge from the source coresets is explicitly copied to the target coreset only if it expands the coverage of the nominal zone.

\section{Results}
\label{sec:results}

\subsection{Datasets}
We introduce a dual dataset of $10\text{cm}\times12\text{cm}$ aluminum plates containing defects and impurities.
The real dataset is acquired by a visual inspection system, and the synthetic one is generated with the inspection context modeled to match the real system.
The plates are flat, uncoated, and exhibit slight brushed texture.
The simple geometry was chosen to remove inspection difficulties arising from complex geometry and allow focusing on defect recognition only.
The real dataset contains $18$ sample plates, with $7$ of them containing bumps, dents and scratch defects on both sides of the plates.
Impurities in form of water stains, fingerprints and numbered stickers were added to only one side of the defective samples.
Both defects and impurities were manually annotated using LabelMe \cite{WadaLabelMe}.
The samples are inspected from both sides using a robotic manipulator carrying the camera within a ring light source.
The inspection setup is equivalent to the setup presented in \cite{Fulir2023}.
For each sample $12$ images of size $2448\times 2050$ are acquired from a grid of locations perpendicular to the surface.
We mask out the plate borders and background to remove the influence of geometrical imperfections inflicted by the cutting process and allow focusing on the evaluation of impurity influence.
We use $10$ clean samples for training, $1$ defected sample for threshold estimation and $1$ clean with $6$ defective for testing.

We generate the synthetic training data containing correct samples using physically-based path tracing following the pipeline presented in \cite{Bosnar2020pipeline}.
The surface texture model is adapted from the parallel texture model \cite{Bosnar2022texture}, using custom parameters manually set to mimic the real images.
We produce two versions of the dataset, one containing only clean surfaces and another containing water stains modeled in \cref{sec:water_stain}.
For both versions we generate $80$ samples with texture parameters randomly varied, rendering one image in the middle of the plate for each.
The simulation of masked-out plate borders is moved to data augmentation step.
The texture parameters realizations have been kept exactly the same for both datasets to ensure the only difference is the addition of water stains.

Synthetic data is crucial for the comparison since it is almost impossible to guarantee the reproducibility of real data surface appearance before and after adding water stains. 
This inconsistency arises due to environment factors such as surface handling artifacts, acquisition imprecision and the reactivity of the sample material.
Reproducibility allows us to filter out the differences resulting merely from the difference in texture.
The domain randomization strategy follows that of Wagenstetter \etal ~\cite{Wagenstetter2024}, which has shown to be successful for anomaly detection on flat metal surface.
Two additional datasets are rendered with perturbed light source: rotated by $90\deg$, or scaled up $2$ times.
This produces different reflections close to the edges of reflected light source and increases the overexposed area (Appendix \ref{apdx:domain_randomization_examples}).

\subsection{Experiment setup}
In all experiments we use our sequential version of PatchCore, reworked from the original implementation available in \textit{anomalib} library \cite{akcay2022anomalib}.
For coreset size we use $2048$ with only $1$ nearest neighbor.
We found the feature extractor of MobileNetv3 size \textit{large} to provide the best speed to performance ratio compared to other ImageNet \cite{ImageNet} pre-trained MobileNetv3 \cite{Howard2019mobilenetv3}, WideResNet \cite{Zagoruyko2016wideresnet}, ResNet \cite{He2016resnet} and VGG \cite{Simonyan2015vgg} architectures.
We construct our feature vectors by concatenating the 3rd and 4th layers as described in \cite{Roth2022patchcore}.
However we had no performance benefit of using the random sparse projection thus we removed it from the experiments.
We average pool the feature maps using kernel size $2$ and blur the resulting anomaly maps with Gaussian kernel of standard deviation $2$ and kernel size $16$.
The original deviation of $4$ has shown to be overly sensitive in the overexposed areas of the image, producing excessive false-positives.
We maximized the chunk size to $2048$ to fit the calculations on a single device per experiment, a Nvidia TITAN V graphical processing card with $12GB$ of memory.
After collecting the coreset, we threshold the resulting anomaly maps by maximizing the $F_1$ score on the validation split of real data.

For experiments where we explicitly use data augmentation (DA), we randomly increase brightness and contrast, flip along both axes, and apply Gaussian blur and noise. 
Since the synthetic datasets were rendered to contain the entire plane, we always randomly mask out a region of the image to simulate the rapid transition from surface texture into the black background.

The coreset is collected on the train split containing only correct images and the anomaly threshold is then estimated using the validation split of the defected real dataset to perform binary anomaly segmentation.
Training over the entire trainset is performed for $5$ epochs.
Two types of synthetic data are used: without water stains (Synth) and with water stains (Synth WS).
Domain randomization (DR) experiments use a trainset made from concatenation of the original dataset with the dataset versions where the light source is either rotated by $90\deg$ or upscaled $2$ times.
Data augmentation (DA) experiments apply previously described data augmentation to the basic trainset.
Combined data augmentation and domain randomization (DA+DR) apply augmentation to the concatenated domain randomization trainset.
Finetuning experiments (ft) use the pre-trained coreset and update it using the real trainset.
Coreset melding is formed from coresets trained separately on real, synthetic, domain randomized (DR) and data augmented (DA) trainsets reported in table \cref{tab:trainingroutines}.

The experiments are evaluated on the test split of the real dataset.
The binary pixel-wise precision (P$_{\text{PX}}$), recall (R$_{\text{PX}}$) and F$_1$ score (F$_{\text{1,PX}}$) are reported to show the model performance in the task of covering the defects.
We also measure the pixel-wise per-class coverage for each class to assess their influence on the model predictions.

In surface inspection, it is important to know how many of the defects were detected, while precise defect segmentation is required less often.
Therefore, we introduce the defect-wise recall ($R_{DW}$) which measures the ratio of defects detected by anomaly segmentation to the total number of defects, with defects being counted unique in each image using defect instance masks.
We summarize the mean defect-wise recall ($mR_{DW}$) averaged over all defect classes.

\subsection{Influence of impurities}
\label{sec:results_imperfections}

\begin{table*}
    \caption{Per-class pixel-wise and defect-wise recall values [\%], with and without finetuning (ft).}
    \centering
    \begin{tabular}{l|rr|rr|rr||rr|rr|rr|rr}
        & \multicolumn{2}{c}{Scratch$_\uparrow$} & \multicolumn{2}{c}{Bump$_\uparrow$} & \multicolumn{2}{c}{Dent$_\uparrow$} & \multicolumn{2}{c}{WaterStain$_\downarrow$} & \multicolumn{2}{c}{Fingerprint$_\downarrow$} & \multicolumn{2}{c}{Sticker$_\downarrow$}   \\
        Trainset               & $R_{\text{PX}}$ & $R_\text{DW}$        & $R_{\text{PX}}$ & $R_\text{DW}$     & $R_{\text{PX}}$ & $R_\text{DW}$     & $R_{\text{PX}}$ & $R_\text{DW}$             & $R_{\text{PX}}$ & $R_\text{DW}$     & $R_{\text{PX}}$ & $R_\text{DW}$ \\ 
        \cline{1-13}
        Real                   & 80.5            & \textbf{99.0}        & \textbf{49.7}   & \textbf{100.0}    & \textbf{96.0}   & 97.7              & 23.5            & 75.9                      & 4.3             & 93.8              & 34.0            & 100.0 \\
        Synth                  & \textbf{82.6}   & 97.5                 & 33.7            & 99.4              & 93.1            & \textbf{98.5}     & 15.3            & 52.5                      & \textbf{1.2}    & 79.2              & 30.5            & 100.0 \\
        Synth$_\text{WS}$      & 72.9            & 95.0                 & 31.4            & \textbf{100.0}    & 85.3            & 97.0              & 5.3             & \textbf{37.9}             & \textbf{1.1}    & 75.0              & 23.4            & 100.0 \\
        Synth (ft)             & 75.5            & 96.5                 & 45.3            & \textbf{100.0}    & 94.3            & 95.5              & 17.6            & 66.5                      & 2.1             & 89.6              & 29.0            & 100.0 \\
        Synth$_\text{WS}$ (ft) & 68.1            & 93.3                 & 35.4            & \textbf{100.0}    & 91.0            & 93.2              & 11.6            & 51.6                      & \textbf{1.0}    & 79.2              & 25.8            & \textbf{96.8}         
    \end{tabular}
    \label{tab:multiclass}
\end{table*}
The influence of different trainsets can be compared and analyzed using the results in \cref{tab:multiclass}.
There, we show the summary of recall values for both pixel-wise (PX) segmentation and defect-wise (DW) detection.
The values measure the percentage of labeled pixels predicted as anomalies and the percentage of instances detected by at least a single prediction.
The measurement is a rough estimate since the imprecise predictions might leak onto overlapping classes.
However, it is sufficient for comparison of relative influences.

First we compare the measured per-class recall with visual examination of predictions.
That way it is possible to analyze the influence of individual classes within defect and impurity categories and identify the common causes of miss-classifications.
The low pixel-wise recall over \textbf{bumps} emphasizes the mismatch between manually annotated labels and what model is actually capable to predict.
The model rarely covers their ground truth completely due to their similarity to the surface texture, leading to ambiguous definition of their borders.
The ambiguity was also noticed during manual annotation (Appendix \ref{apdx:performance_degradation}).
However, the high defect-wise recall of bump detection shows that they contain regions that are easy to detect, likely due to the sharp shadows within them.
\textbf{Dents} show to be easy to cover and detect due to the very distinct edges and inner patterns.
Covering \textbf{scratches}, on the other hand, shows to be somewhat harder, as can be observed in the lower pixel-wise recall values.
This is due to the influence of low contrast in some regions of the scratch, as it often stretches over surfaces causing the illumination transition from bright to dark-field.
\textbf{Fingerprints} are mostly uncovered, likely due to their low contrast.
However they are often detected since they cover a larger area, increasing the probability of intersection with other defects or impurities.
Their low coverage opens the opportunity for filtering them out based on prediction size (Appendix \ref{apdx:prediction_filtering}).
\textbf{Stickers} produce false-positives only along their edges, edges of the marker and bubbles underneath, due to the unexpected sharp edges around them.
Nonetheless, its irregular edges take up a lot of pixels increasing recall value.
\textbf{Water stains} are easily detected and moderately covered by predictions.
They produce false-positives primarily near the high-contrast edges around the overexposed areas or when close to the edges of other defects forcing the model to mistake them as a part of a defect.

When comparing the performance between trainsets, we can observe that the Real and Synth perform much better in terms of defect detection, compared to Synth$_{\text{WS}}$.
We can observe that the recall is generally lowered for Synth$_{\text{WS}}$.
This is likely due to the influence of feature similarity between water stains and defects, which would cause the model to have difficulty deciding if the surface is anomalous or not, and lead to the increase in anomaly threshold.
Apart from the scratches, finetunning moves recall values towards the values obtained on the Real dataset since the synthetic patches in the coreset are being gradually replaced by real patches.

\subsection{Comparison of different training approaches}
\label{sec:results_traincompare}

\begin{table}[t]
    \caption{
        Comparison of different techniques for transfer learning [\%], trained using real data (Real), synthetic data (Synth) or synthetic water stains data (Synth$_{\text{WS}}$).
        Some experiments use modified data using data augmentation (DA) or domain randomization (DR) or both (DR+DA) or they were fine-tuned on real data (ft).
    }
    \centering
    \begin{tabular}{l|rrr|r}
        Training scenario & P$_{\text{PX}}$ & R$_{\text{PX}}$ & F$_{\text{1,PX}}$ & mR$_{\text{DW}}$ \\
        \hline
        Real                & 35.7 & \textbf{68.4} & 47.0 & \textbf{98.9} \\
        \hline
        Synth               & 26.1 & 62.7 & 36.8 & 98.4 \\
        Synth$_{\text{DA}}$            & 31.2 & 60.8 & 41.2 & 98.4 \\
        Synth$_{\text{DR}}$            & 29.4 & 66.7 & 40.8 & \textbf{99.4} \\
        Synth$_{\text{DR+DA}}$        & 27.4 & 66.2 & 38.7 & \textbf{99.2} \\
        \hline
        Synth$_{\text{WS}}$             & 23.3 & 56.2 & 32.9 & 97.3 \\
        Synth$_{\text{WS+DA}}$          & 27.7 & 58.0 & 37.5 & 97.9 \\
        Synth$_{\text{WS+DR}}$          & 24.4 & 59.7 & 34.6 & 98.4 \\
        Synth$_{\text{WS+DR+DA}}$      & 29.4 & 52.0 & 37.5 & 95.3 \\
        \hline
        Synth (ft)          & 41.6 & 63.8 & 50.3 & 97.3 \\
        Synth$_{\text{DA}}$ (ft)       & 45.6 & 57.9 & 51.0 & 96.1 \\
        Synth$_{\text{DR}}$ (ft)       & 41.3 & 63.7 & 50.1 & 96.5 \\
        Synth$_{\text{DR+DA}}$ (ft)   & 47.3 & 58.2 & \textbf{52.2} & 95.2 \\
        \hline
        Synth$_{\text{WS}}$ (ft)        & 50.3 & 55.6 & \textbf{52.9} & 95.5 \\
        Synth$_{\text{WS+DA}}$ (ft)     & 49.5 & 55.8 & \textbf{52.5} & 95.0 \\
        Synth$_{\text{WS+DR}}$ (ft)     & 44.5 & 58.3 & 50.5 & 95.9 \\
        Synth$_{\text{WS+DR+DA}}$ (ft) & 44.5 & 55.7 & 49.5 & 95.6 \\
        \hline
        Melding             & 51.5 & 52.6 & \textbf{52.0} & 93.9 \\
        Melding$_{\text{WS}}$          & \textbf{53.9} & 50.1 & \textbf{51.9} & 91.7
    \end{tabular}
    \label{tab:trainingroutines}
\end{table}

In \cref{tab:trainingroutines} we compare the binary anomaly segmentation and mean defect-wise anomaly detection result results of different training approaches.
We first observe that the overall recall values are low, which is mainly hindered by the annotation ambiguity of bumps as observed in \cref{sec:results_imperfections}.
Synthetic data on its own performs well but not as good as 
the real dataset which is due to the domain difference affecting the mapping of the feature extractor.
Between the two synthetic dataset versions, Synth$_{\text{WS}}$ produces worse results even though it simulates the water stains present in the real dataset.
This is affected by the perceptual similarity between dents and water stains which increases the recognition difficulty.
However, if we fine-tune the model on real data the situation greatly changes and the models outperform the real data 

Finetuning primarily increases precision, indicating that the zone of nominal samples is rearranged to accommodate for the domain gap.
Melding the pre-trained coresets slightly trades off recall for precision, but achieves an effect similar to finetuning while increasing the speed of training.
This is due to the fact that we can train multiple models in parallel on different machines and meld only their coresets, which are significantly smaller than the trainset.
Since we can control the target size, we additionally tried melding the pre-trained coresets into melded coresets double and half the original size.
However, we have not observed any benefits from either.

Domain augmentation performs superior to data randomization, even though randomization generates samples that are differently illuminated.
This is likely due to the inclusion of blurring in the synthetic images through augmentation, which more closely simulates the blurring resulting from the camera lens.

\section{Discussion}
\label{sec:discusstion}
Extending the synthetic data with water stains has shown to be beneficial and help lowering the water stain recall significantly (see \cref{tab:multiclass}).
This indicates that the realism of the introduced water stain model was sufficient for the purpose of this work.
However, it is important to note that the introduced model generates water stains with exponential decay in reflectance, which might not always be the case in real water stains. 
As can be seen in see \cref{fig:water_stains}, real water stains can also exhibit heterogenity within the outline.
For that purpose, it is advisable to extend the model in the future to allow this kind of control.

Consistent with previous studies \cite{Fulir2023,Wagenstetter2024}, we observed a performance decline when using only synthetic data.
This is caused by the domain gap and finetuning helps us bridge this gap and further exploit the information from synthetic data.
The performance becomes much better than using solely real data and the defect-wise recall shows that the majority of defects were found.

Further research is needed to describe how the domain gap is formed by the coresets and which tools can be used to close it.
Additionally, similarity between dents and water stains introduces false-positives which significantly reduce precision.
The likely source of issues in both cases is the feature extractor which might not be extracting features that are general enough to ignore inter-domain differences and not specialized enough to recognize the unique features between the classes \cite{Cui2023unsupervised_ad_survey}.
Another source is the coreset building algorithm itself: which samples it chooses and how the final threshold is estimated based on them.
Adjusting the feature extractor to the features descriptive of the target domain should help resolve these issues.

Defect-wise recall provided a different view.
Compared to pixel-wise recall, defect-wise recall shows that the models in fact find a high percentage.
The defect-wise metrics were able to overcome the ambiguity of annotating the bumps and appearance changes of scratches.
The bumps contain regions which very gradually move away from the nominal surface texture, increasing the difficulty to defining its exact borders.
The problem was resolved by ignoring the exact defect coverage and focusing on the number of defect detections.
It is however difficult to define how much coverage can be considered a detection, since model predictions can leak from nearby defects and only slightly cover a nearby defect.
We reported the one-pixel detection results since we observed that filtering out small predictions lowers the performance over defect classes (Appendix \ref{apdx:prediction_filtering}).
However, increasing the model precision would resolve this issue.

PatchCore \cite{Roth2022patchcore} suffers from inherent memory cost due to the need of explicit storage of nominal feature vectors for the entire trainset.
Our implementation replaces the memory cost with multiple iterations through the trainset.
It can chunk the number of patches and adjust the memory cost to the device, allowing us to run it on a GPU, which massively parallelizes the execution and significantly speeds up the calculation of patch scores.

In addition to the memory benefits, coreset melding enabled us to efficiently meld multiple coresets into a single refined one, where the final model complexity can be controlled by the size of the coreset.
Coreset melding is a form of federated learning \cite{Li2023federated_survey} since it parallelizes the training over different datasets and devices before merging them into a single model.
The algorithm might also be beneficial for continual learning since the coreset building criterion prioritizes expansion of the nominal zone \cite{Li2022continual_ad}.
Additional research might find this method useful for various applications outside of the industrial inspection domain.

\section{Conclusion}
\label{sec:conclusion}

We introduce Sequential PatchCore which resolves the known memory bottleneck of PatchCore on high resolution images.
In addition, we introduce the novel coreset melding approach which increases the speed of transfer learning through parallelized training, while achieving competitive performance.

In this work, we have, for the first time, evaluated the influence of water stains on anomaly detection.
The reproducibility of the synthetic data generation made the evaluation possible by allowing us to measure only the influence of the water stains.
Synth impurities helped simulate the features missing in train data, while keeping all the other parameters the same, unaffected by the changes present in the real environment.
The realism of the proposed water stain model based on Perlin noise and jittered sampling has proven sufficient for reducing the sensitivity of the resulting model on water stains. 

Although the model itself became less sensitive to water stains, training on impurities reduced performance on defects.
This supports the observed behavior of impurities having similar features with defects. 
However, once the the model trained on water stains was fine-tuned, the precision and F$_1$ score increased significantly.
Overall, the pixel-wise metrics still report relatively low performance due to their reliance on pixel-perfect annotations and predictions, which are almost impossible to ensure.
Therefore, the introduced defect-wise recall has provided a new perspective on model performance, pointing out the incomplete conclusions stemming from standard metrics.

\section*{Acknowledgements}
This work was supported by Fraunhofer ITWM. We thank Lovro Bosnar for his valuable feedback and discussions, and Jonas Ruppert for helping prepare the real datasets.

\bibliographystyle{ieeetr}
\bibliography{main}

\clearpage
\appendices

\section{Examples from the datasets}
\label{apdx:domain_randomization_examples}

By altering the rotation and scaling of the light source, we achieve domain randomization to observe changes on the object's surface under different lighting conditions (main paper, Chapter 5.1 Datasets). 
We use three different light sources: a hexagonal light consistent with the real-world light, the same light rotated 90 degrees along the Z-axis, and the same light doubled in size. 
While the range of variations in texture and water stains is the same across the three light sources, the specific parameters generated are different, making each image unique.

\begin{figure}[b!]
    \centering
    \begin{subfigure}[t]{0.32\linewidth}
        \centering
        \includegraphics[width=\linewidth]{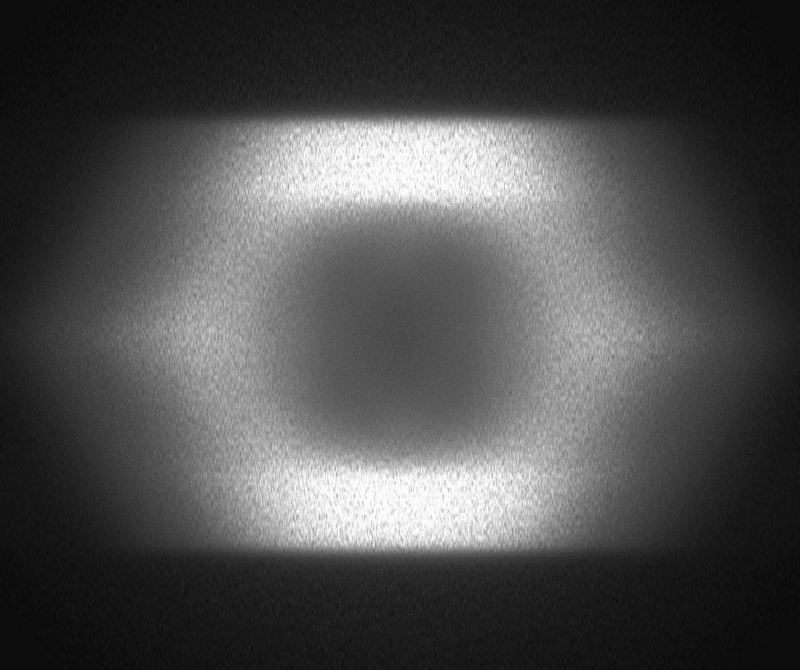}
    \end{subfigure}
    \begin{subfigure}[t]{0.32\linewidth}
        \centering
        \includegraphics[width=\linewidth]{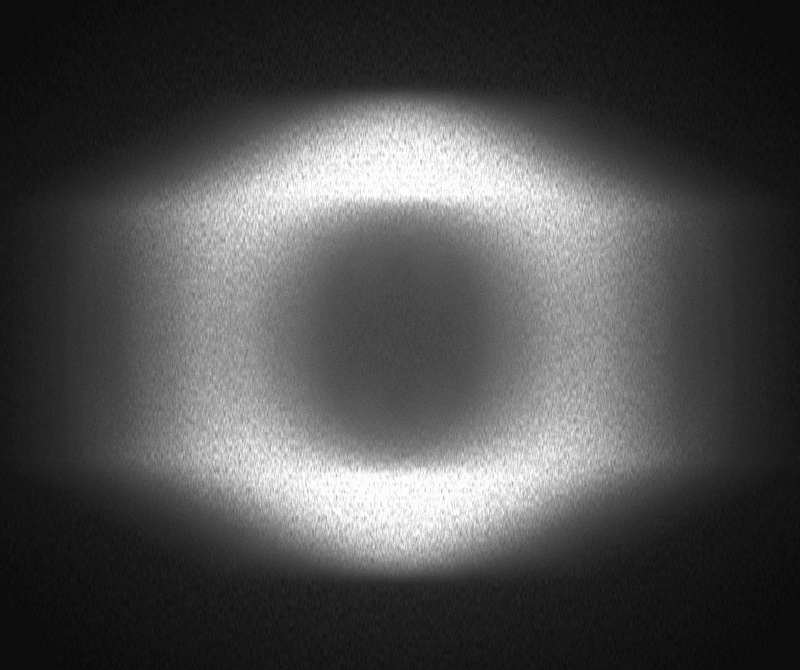}
    \end{subfigure}
    \begin{subfigure}[t]{0.32\linewidth}
        \centering
        \includegraphics[width=\linewidth]{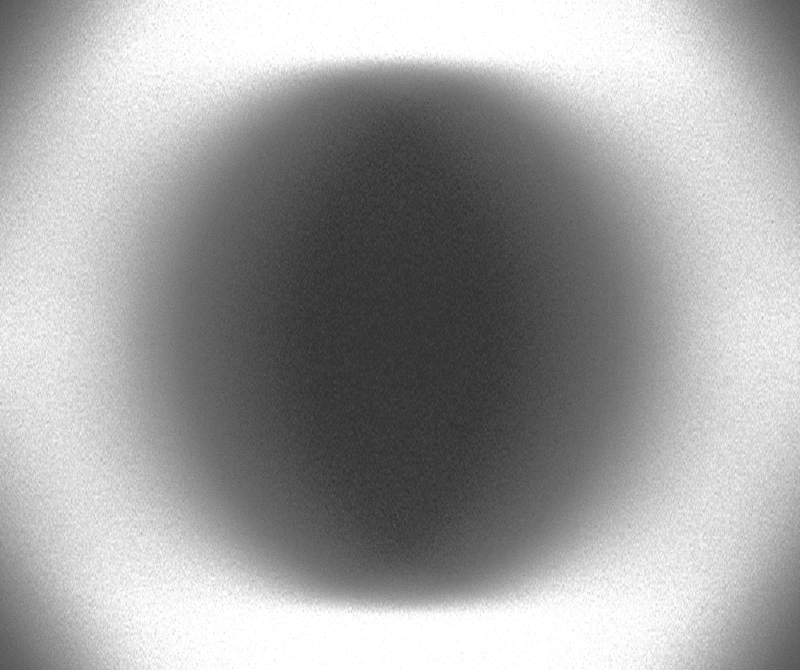}
    \end{subfigure}
    \hspace{0.02\linewidth}
    \begin{subfigure}[t]{0.32\linewidth}
        \centering
        \includegraphics[width=\linewidth]{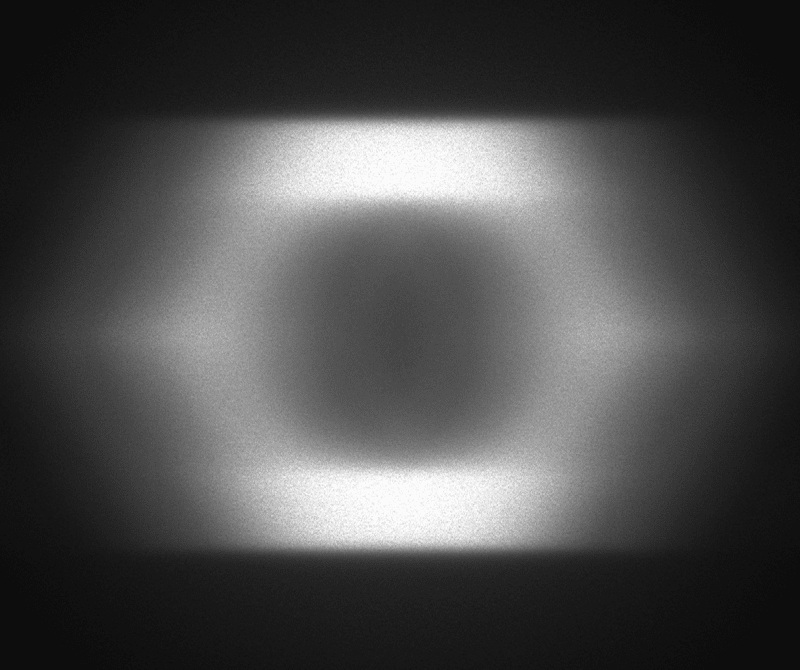}
    \end{subfigure}
    \begin{subfigure}[t]{0.32\linewidth}
        \centering
        \includegraphics[width=\linewidth]{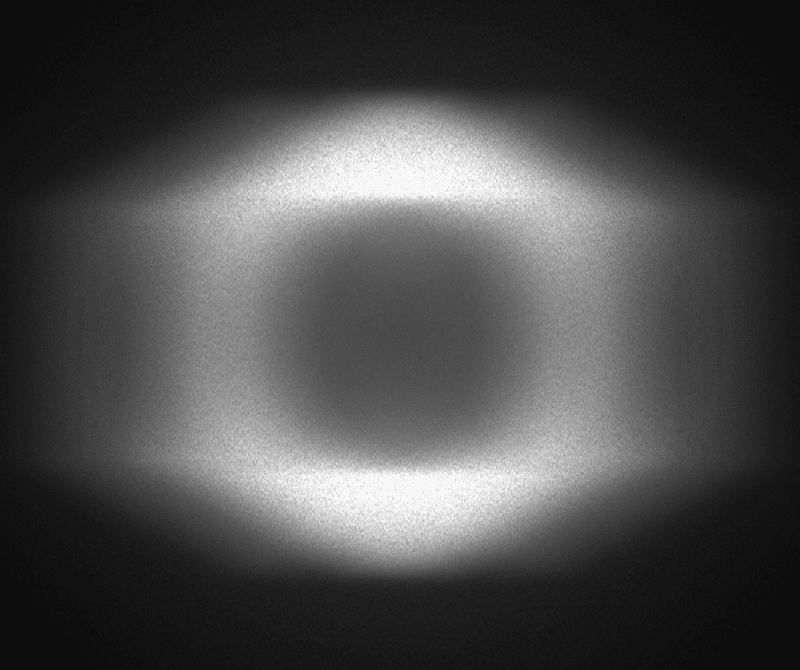}
    \end{subfigure}
    \begin{subfigure}[t]{0.32\linewidth}
        \centering
        \includegraphics[width=\linewidth]{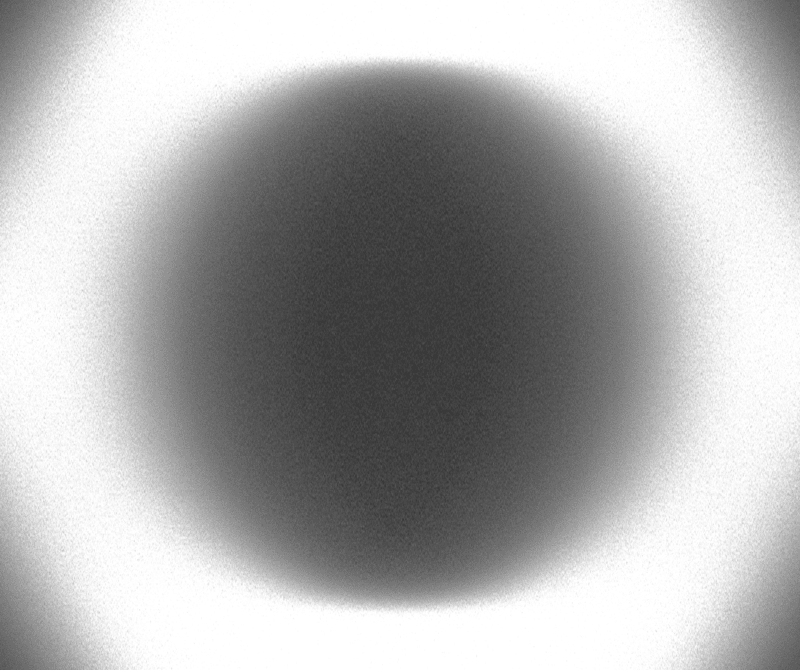}
    \end{subfigure}

    \vspace{0.1cm}
    
    \begin{subfigure}[t]{0.32\linewidth}
        \centering
        \includegraphics[width=\linewidth]{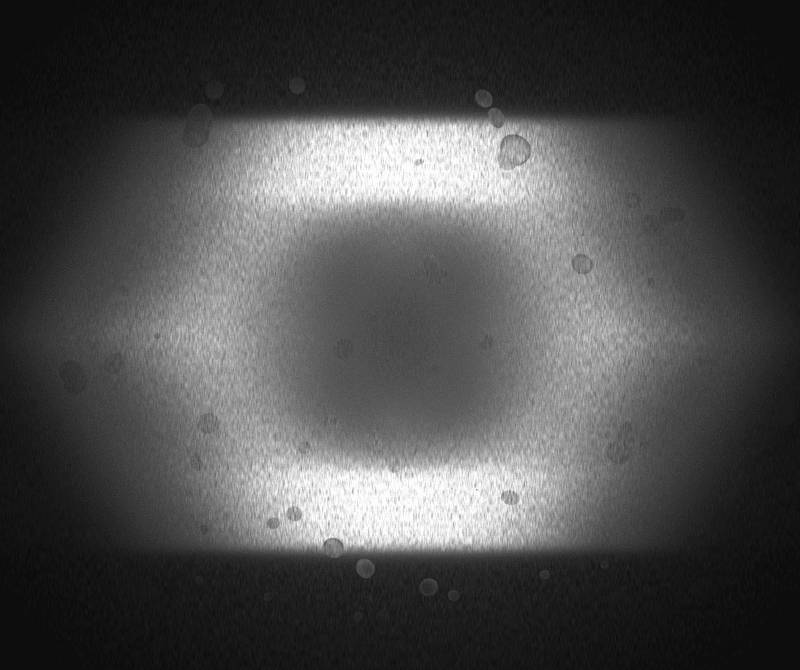}
    \end{subfigure}
    \begin{subfigure}[t]{0.32\linewidth}
        \centering
        \includegraphics[width=\linewidth]{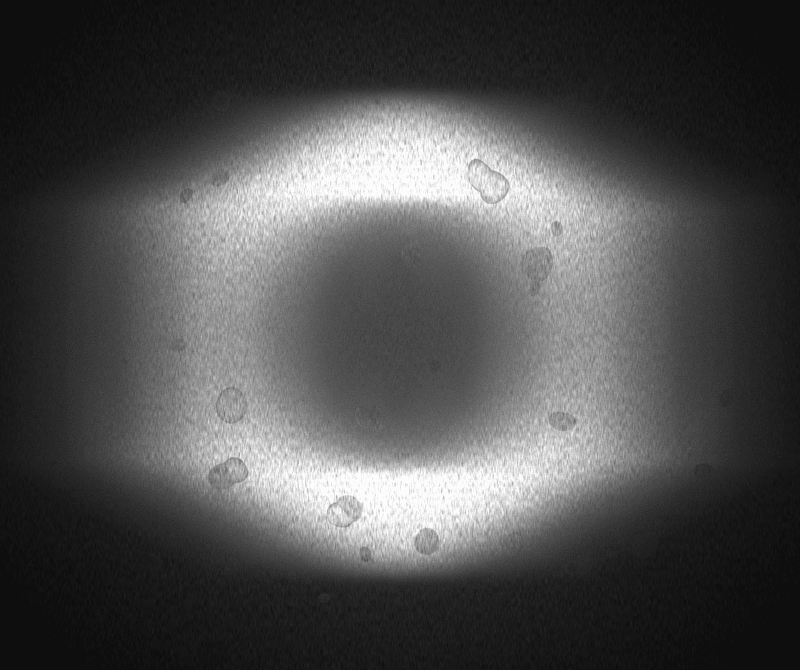}
    \end{subfigure}
    \begin{subfigure}[t]{0.32\linewidth}
        \centering
        \includegraphics[width=\linewidth]{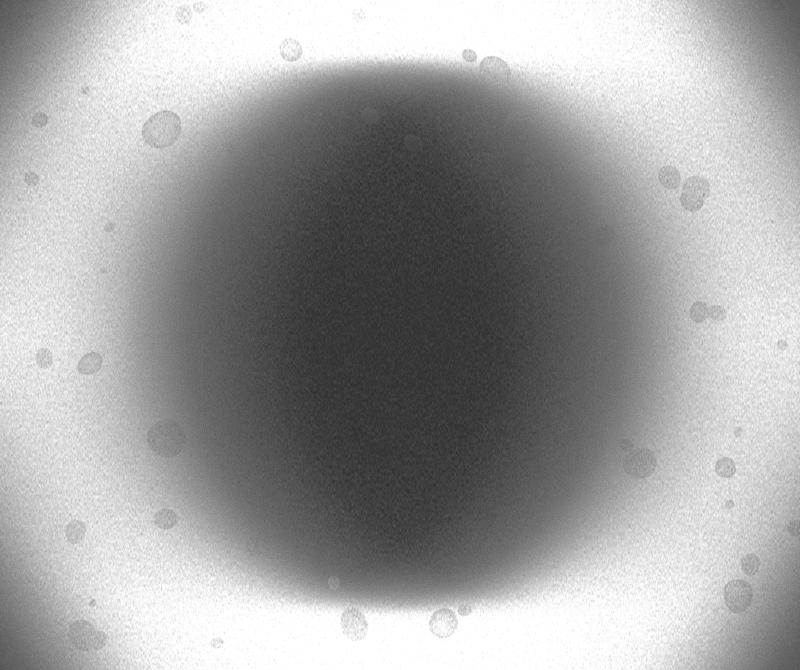}
    \end{subfigure}
    \hspace{0.02\linewidth}
    \begin{subfigure}[t]{0.32\linewidth}
        \centering
        \includegraphics[width=\linewidth]{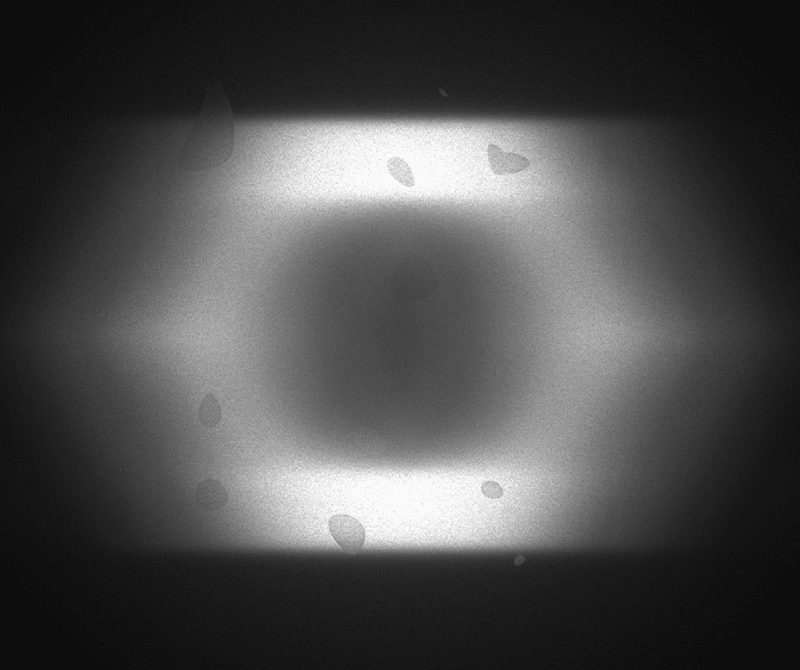}
    \end{subfigure}
    \begin{subfigure}[t]{0.32\linewidth}
        \centering
        \includegraphics[width=\linewidth]{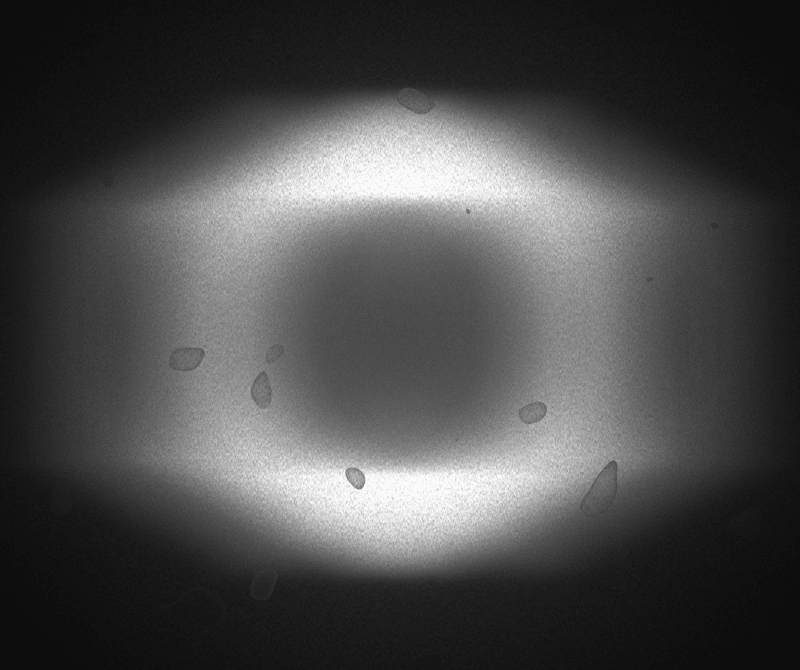}
    \end{subfigure}
    \begin{subfigure}[t]{0.32\linewidth}
        \centering
        \includegraphics[width=\linewidth]{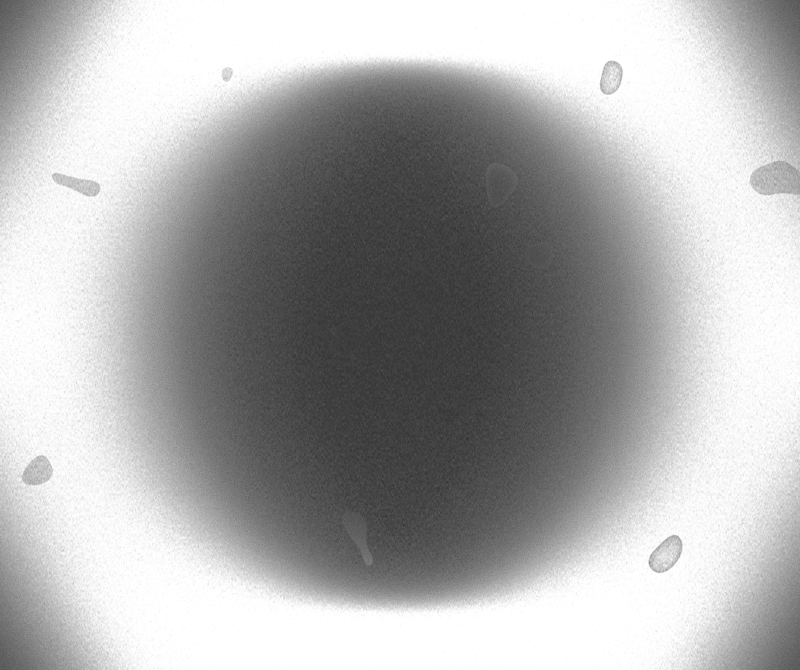}
    \end{subfigure}

    \vspace{0.1cm}
    
    \begin{subfigure}[t]{0.32\linewidth}
        \centering
        \includegraphics[width=\linewidth]{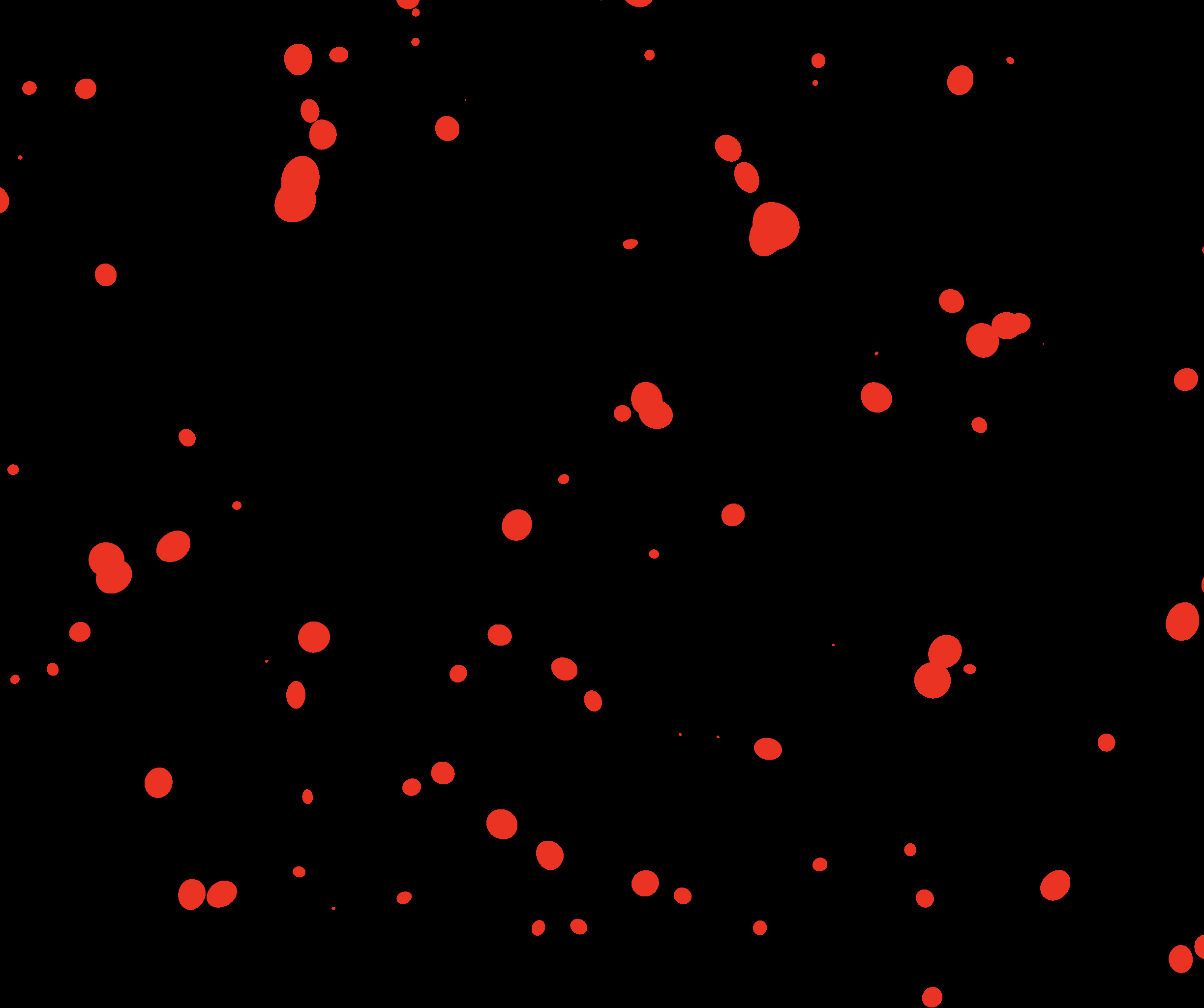}
    \end{subfigure}
    \begin{subfigure}[t]{0.32\linewidth}
        \centering
        \includegraphics[width=\linewidth]{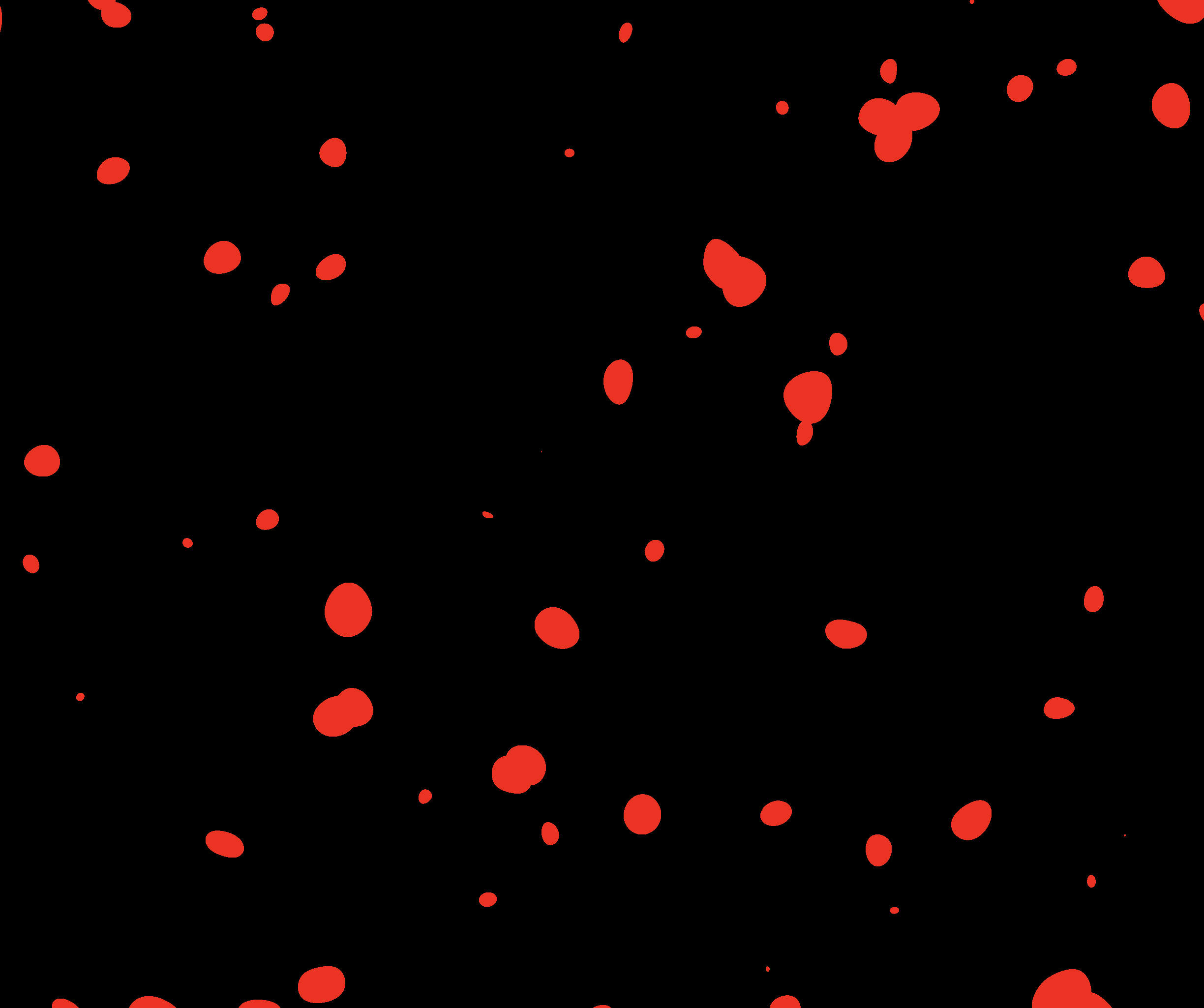}
    \end{subfigure}
    \begin{subfigure}[t]{0.32\linewidth}
        \centering
        \includegraphics[width=\linewidth]{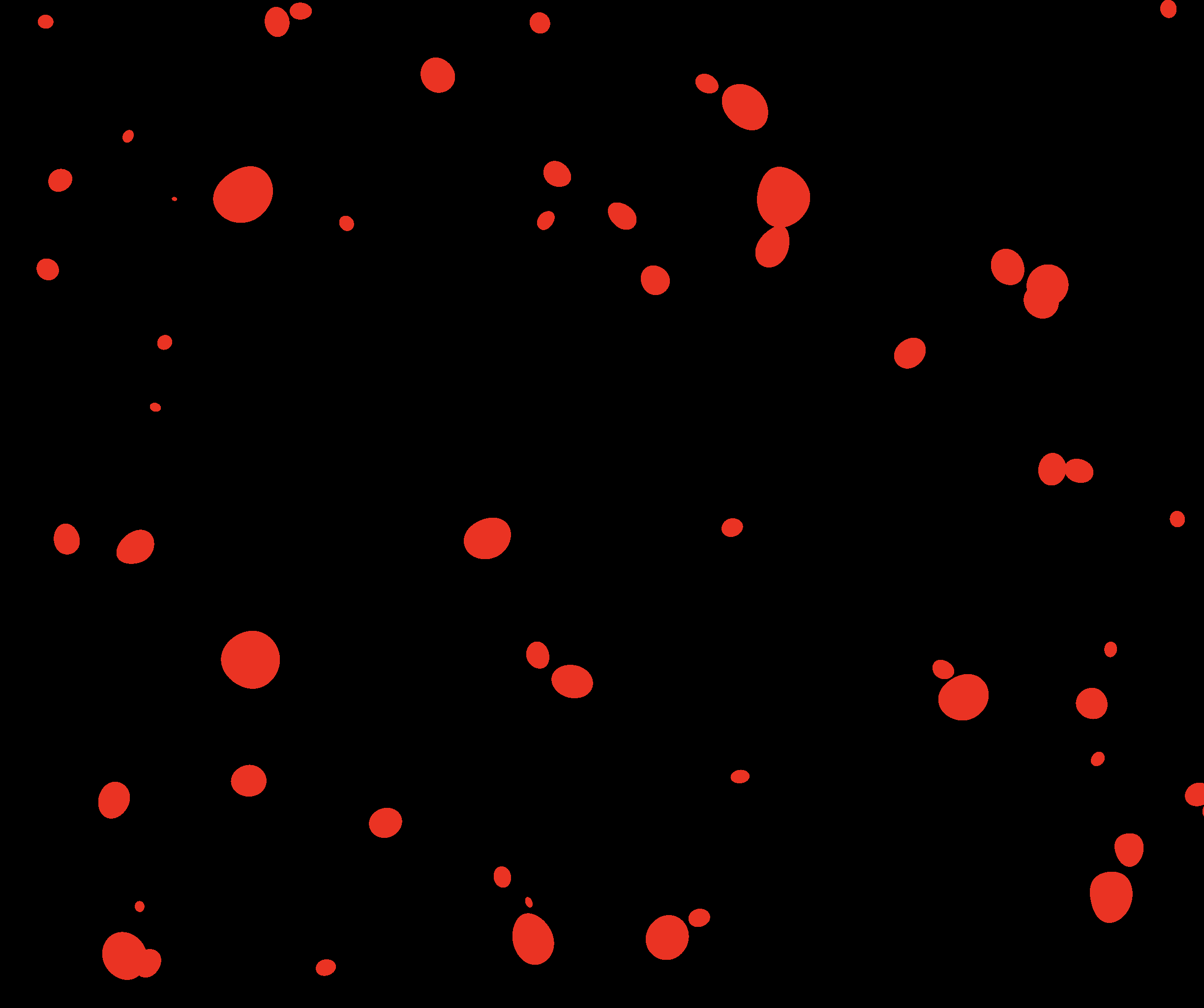}
    \end{subfigure}
    \hspace{0.02\linewidth}
    \begin{subfigure}[t]{0.32\linewidth}
        \centering
        \includegraphics[width=\linewidth]{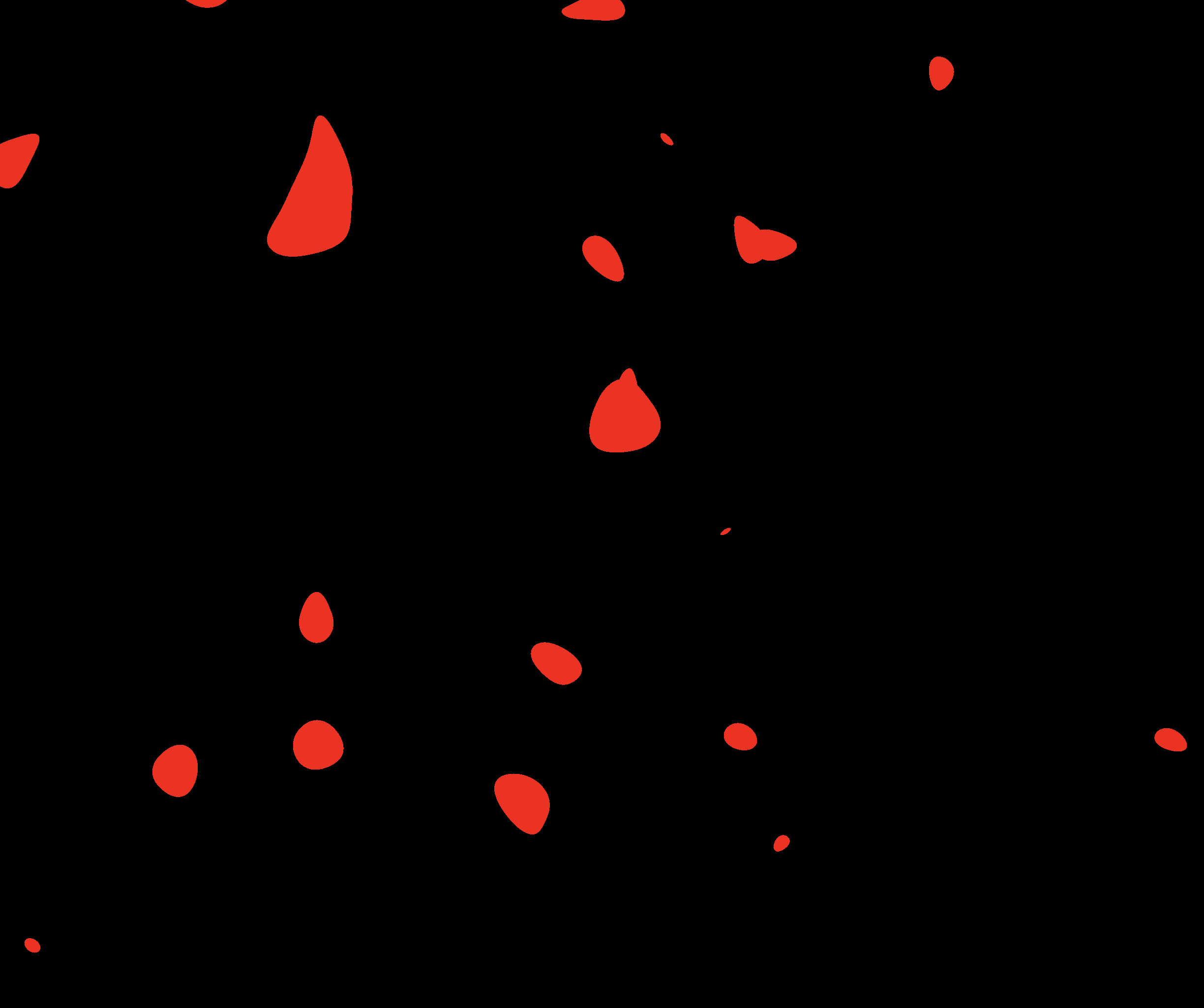}
    \end{subfigure}
    \begin{subfigure}[t]{0.32\linewidth}
        \centering
        \includegraphics[width=\linewidth]{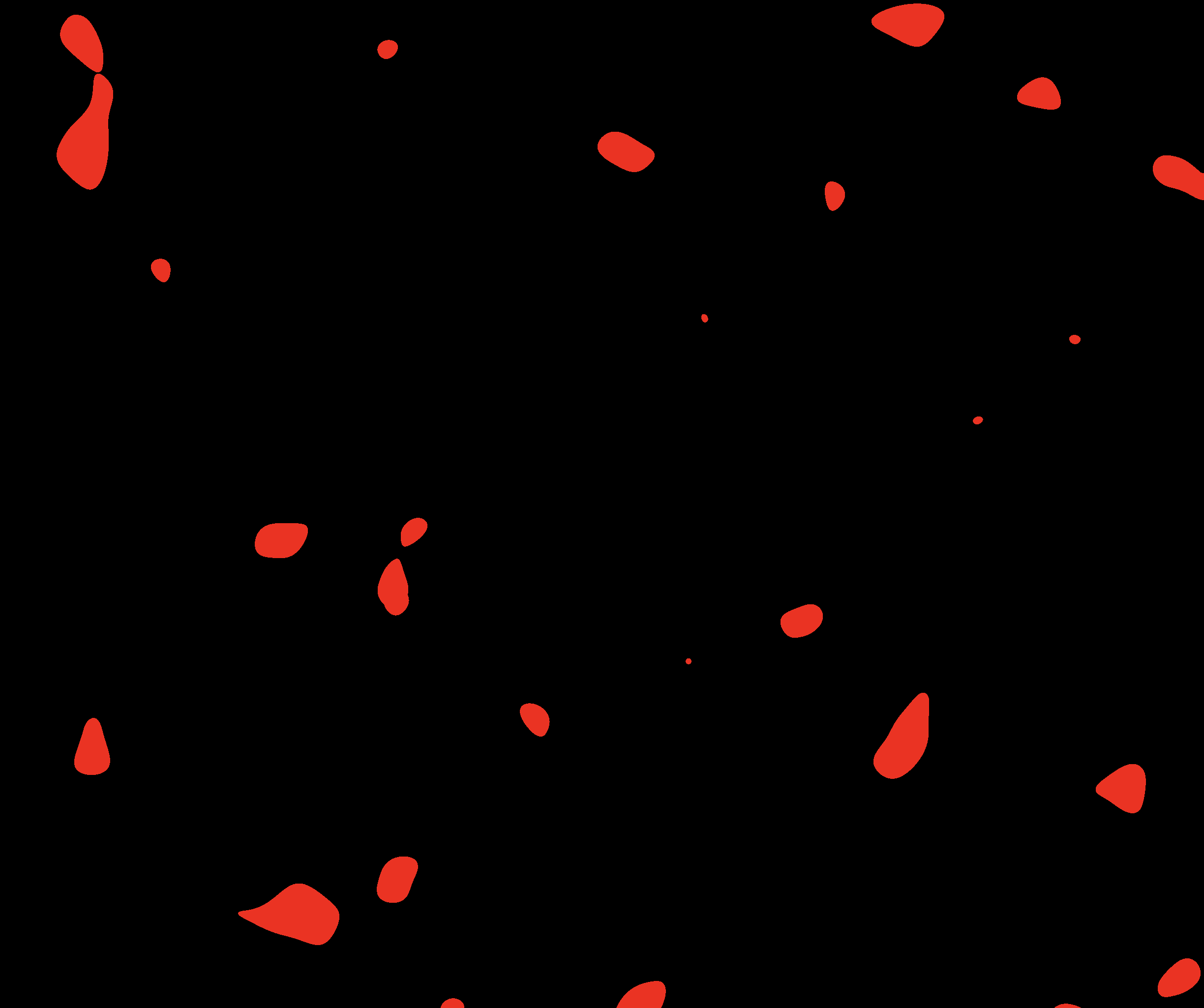}
    \end{subfigure}
    \begin{subfigure}[t]{0.32\linewidth}
        \centering
        \includegraphics[width=\linewidth]{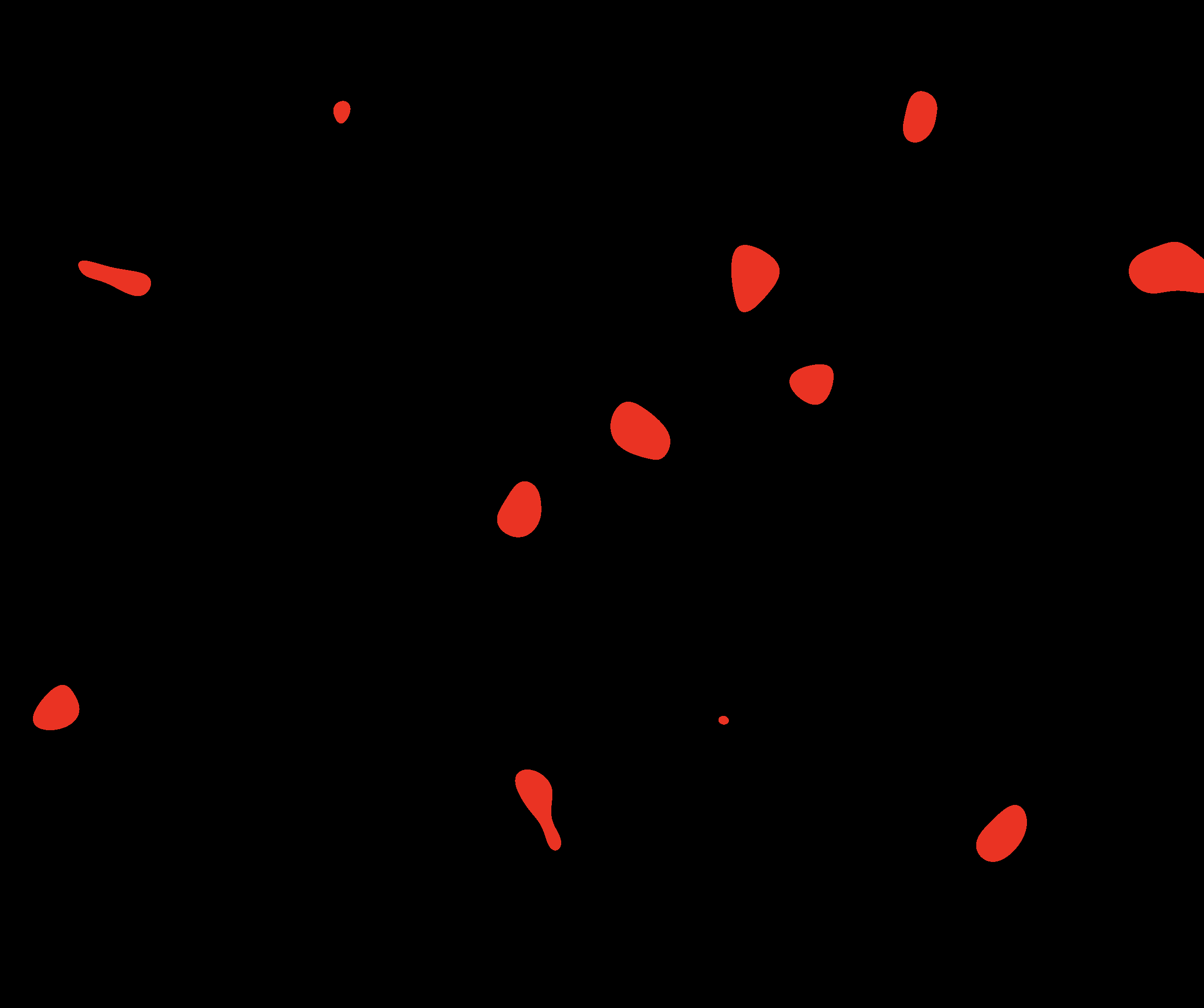}
    \end{subfigure}

    \caption{Differences in two groups of texture and water stain distribution patterns under different lighting conditions. The first row in each group displays different illumination patterns, the second row shows the effects of adding water stains, and the third row presents water stain masks on a black background, with red areas indicating water stains. The images on the left exhibit one texture and water stain distribution pattern, while the images on the right show another pattern.}
    \label{fig:different_illumination}
\end{figure}

In \cref{fig:different_illumination}, we can see how these variations affect the object's surface. 
Without water stains, different lighting conditions alter the brightness and light distribution in the images, creating notable dark regions in the transition areas between the outline of the light source and its surroundings. 
With water stains, different lighting conditions impact the visibility and appearance of the stains. 
In bright areas, the edges of the water stains are more pronounced, while in dark and over-exposed areas, the water stains become blurred and difficult to detect. 
Additionally, the annotations provide a reference for the distribution of water stains.

\section{Common causes performance degradation}
\label{apdx:performance_degradation}

\begin{figure}[b!]
    \centering
    \begin{subfigure}{\linewidth}
        \centering
        \includegraphics[width=0.45\linewidth]{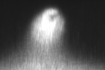}
        \includegraphics[width=0.45\linewidth]{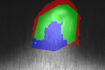}
        \caption{Label bias}
        \label{fig:label_bias}
    \end{subfigure}
    \begin{subfigure}{\linewidth}
        \centering
        \includegraphics[width=0.45\linewidth]{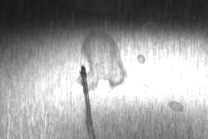}
        \includegraphics[width=0.45\linewidth]{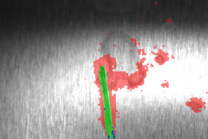}
        \caption{Prediction leakage}
        \label{fig:prediction_leakage}
    \end{subfigure}
    \begin{subfigure}{\linewidth}
        \centering
        \includegraphics[width=0.45\linewidth]{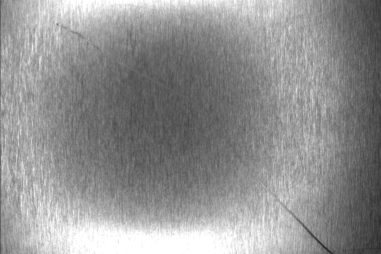}
        \includegraphics[width=0.45\linewidth]{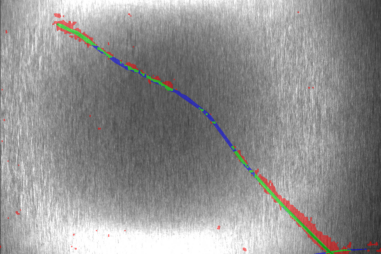}
        \caption{Appearance change}
        \label{fig:appearance_change}
    \end{subfigure}
    \caption{Examples of common effects lowering the model performance. Green pixels are true-positives, red-pixels are false-positives and blue pixels are false-negatives. Best viewed in color.}
    \label{fig:perf_degradation}
\end{figure}

In \cref{fig:perf_degradation} we display examples of effects that lower the model performance across images.
The examples are collected from predictions of the model trained on real data.
The images might be lower resolution than the images in the dataset.
The \textbf{label bias} (\cref{fig:label_bias}) is caused by the difficulty of deciding how to annotate the bump defects.
As seen in the prediction example, model predicted parts of the bump in the shadow area, but missed parts of the near-overexposed light area.
The \textbf{prediction leakage} (\cref{fig:prediction_leakage}) happens when an impurity is nearby a defect, causing the model predictions to "spill" towards the impurity and merge with the false-positive prediction on the impurity. 
The \textbf{appearance change} of long scratches, cause false-negatives where the scratch transitions from bright to dark field and vice-versa.

\section{Filtering out small predictions}
\label{apdx:prediction_filtering}

Small predictions are often an artifact of local patterns triggering the model barely enough to make them overcome the decision boundary.
Usually they present as small isolated regions of false-positives.
When evaluating the recall values (main paper, Tab 1) we observed small coverage but plentiful detection of fingerprints, which indicates possibility of small predictions.
In the industry, this is often filtered out using different filtering mechanisms.

\begin{figure}[h!]
    \centering
    \includegraphics[width=0.75\linewidth]{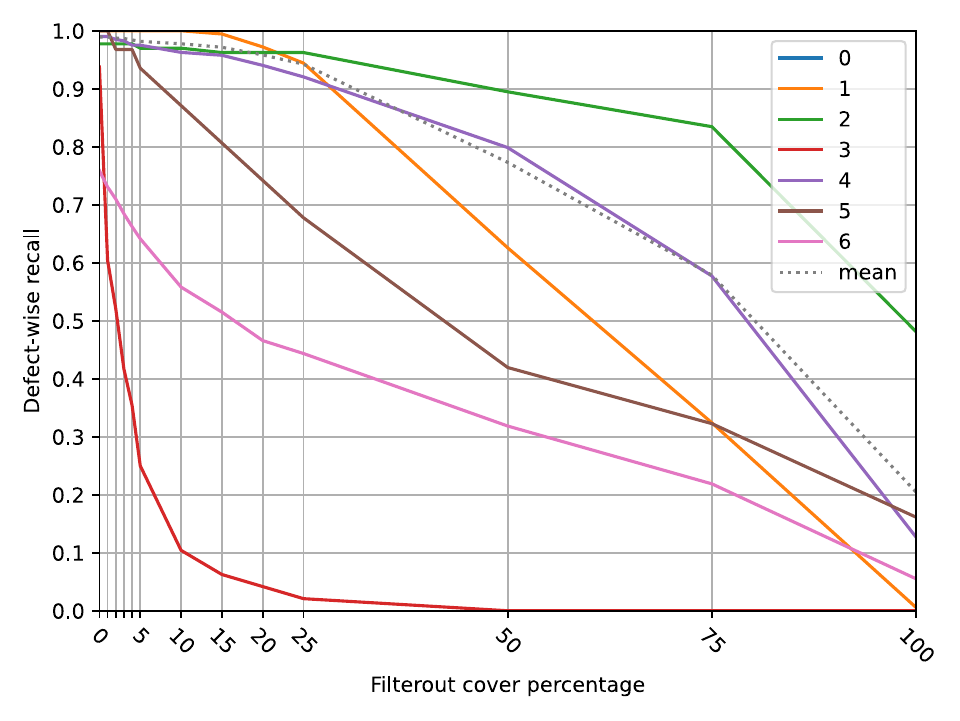}
    \caption{Differences in per-class and mean defect-wise recall depending on the percentage of defect considered as detection. Classes in order are: backgorund, bump, dent, fingerprint, scratch, sticker, water stain}
    \label{fig:prediction_filtering}
\end{figure}

In \cref{fig:prediction_filtering} we present how the defect-wise recall value changes when applying different thresholding for when a prediction is considered a detection.
The results are from a model trained on real data, but graphs on other models are very similar.
For each defect instance we measure the relative defect coverage, in the number of pixels that were predicted divided by the size of defect instance.
The defect coverage is thresholded with different percentages required to consider the coverage as detection.
We consider percentages: 0, 1, 2, 3, 5, 10, 15, 20, 25, 50, 75, 100.

We can see that the impurity classes quickly disappear, especially fingerprints and water stains.
However, this also reduces the coverage of defect classes.
We made the choice to maximize the coverage over defect classes because we follow the interest of avoiding false-negatives over false-positives.
However, depending on the target application the risk of loosing defects might be worth it.

\end{document}